\documentclass{article} %

\usepackage{hyperref}

\usepackage{url}
\usepackage[accepted]{icml2021}

\usepackage{booktabs}
\usepackage{colortbl}

\usepackage{graphicx}
\usepackage{float,wrapfig}

\usepackage{array}
\usepackage{adjustbox}
\usepackage{multirow}
\usepackage{colortbl}

\usepackage{caption}
\usepackage{subcaption}
\usepackage{color,xcolor}
\usepackage{epsfig}
\usepackage{bbm}
\usepackage{hhline}

\usepackage{amsmath,amsfonts,amsthm,amssymb}
\usepackage{mathtools}
\usepackage{bm}
\usepackage{nicefrac}
\usepackage{microtype}

\usepackage{changepage}
\usepackage{extramarks}
\usepackage{fancyhdr}
\usepackage{lastpage}
\usepackage{setspace}
\usepackage{soul}
\usepackage{xspace}

\usepackage{url}
\usepackage{algorithm, algorithmic}
\usepackage{todonotes} %

\usepackage{titlesec}

\usepackage{amsmath,amsfonts,bm}

\def\eqref#1{equation~\ref{#1}}

\def\1{\bm{1}}

\def\vx{{\bm{x}}}

\DeclareMathAlphabet{\mathsfit}{\encodingdefault}{\sfdefault}{m}{sl}
\SetMathAlphabet{\mathsfit}{bold}{\encodingdefault}{\sfdefault}{bx}{n}

\def\gL{{\mathcal{L}}}

\newcommand{\E}{\mathbb{E}}

\newcommand{\R}{\mathbb{R}}

\newcommand{\sect}[1]{Section~\ref{#1}}

\newcommand{\eqn}[1]{Equation~\ref{#1}}
\newcommand{\fig}[1]{Figure~\ref{#1}}
\newcommand{\tbl}[1]{Table~\ref{#1}}
\newcommand{\myparagraph}[1]{{\bf #1}\quad}

\DeclarePairedDelimiterX{\infdivx}[2]{(}{)}{%
  #1\;\delimsize|\delimsize|\;#2%
}
\newcommand{\kldiv}[2]{\ensuremath{\text{KL}\infdivx{#1}{#2}}\xspace}

\definecolor{MyDarkBlue}{rgb}{0,0.08,1}
\definecolor{MyDarkGreen}{rgb}{0.02,0.6,0.02}
\definecolor{MyDarkRed}{rgb}{0.8,0.02,0.02}
\definecolor{MyDarkOrange}{rgb}{0.40,0.2,0.02}
\definecolor{MyPurple}{RGB}{111,0,255}
\definecolor{MyRed}{rgb}{1.0,0.0,0.0}
\definecolor{MyGold}{rgb}{0.75,0.6,0.12}
\definecolor{MyDarkgray}{rgb}{0.66, 0.66, 0.66}

\begin{document}
\twocolumn[
\icmltitle{Improved Contrastive Divergence Training of Energy-Based Model}

\icmlsetsymbol{equal}{*}

\begin{icmlauthorlist}
\icmlauthor{Yilun Du}{mit}
\icmlauthor{Shuang Li}{mit}
\icmlauthor{Joshua Tenenbaum}{mit}
\icmlauthor{Igor Mordatch}{goo}
\end{icmlauthorlist}

\icmlaffiliation{mit}{MIT CSAIL}
\icmlaffiliation{goo}{Google Brain}

\icmlcorrespondingauthor{Yilun Du}{yilundu@mit.edu}

\icmlkeywords{Machine Learning, ICML}

\vskip 0.3in
]

\printAffiliationsAndNotice{} %

\begin{abstract}

 Contrastive divergence is a popular method of training energy-based models, but is known to have difficulties with training stability. We propose an adaptation to improve contrastive divergence training by scrutinizing a gradient term that is difficult to calculate and is often left out for convenience. We show that this gradient term is numerically significant and in practice is important to avoid training instabilities, while being tractable to estimate. We further highlight how data augmentation and multi-scale processing can be used to improve model robustness and generation quality. Finally, we empirically evaluate stability of model architectures and show improved performance on a host of benchmarks and use cases,such as image generation, OOD detection, and compositional generation.

\end{abstract}

\section{Introduction}

Energy-Based models (EBMs) have received an influx of interest recently and have been applied to realistic image generation \citep{han2019divergence, du2019implicit}, 3D shapes synthesis \citep{xie2018learning} , out of distribution and adversarial robustness \citep{lee2018wasserstein, du2019implicit, grathwohl2019your}, compositional generation \citep{hinton1999products,du2020compositional}, memory modeling \citep{bartunov2019meta}, text generation \citep{deng2020residual}, video generation \citep{xie2018video},   reinforcement learning \citep{haarnoja2017reinforcement,du2019model}, continual learning \citep{li2020energy}, protein design and folding \citep{ingraham2019learning,du2020energy} and biologically-plausible training \citep{scellier2017equilibrium}.
Contrastive divergence is a popular and elegant procedure for training EBMs proposed by \citep{Hinton2002Training}  which lowers the energy of the training data and raises the energy of the sampled confabulations generated by the model. The model confabulations are generated via an MCMC process (commonly Gibbs sampling or Langevin dynamics), leveraging the extensive body of research on sampling and stochastic optimization. The appeal of contrastive divergence is its simplicity and extensibility. It does not require training additional auxiliary networks \citep{kim2016deep, dai2019exponential} (which introduce additional tuning and balancing demands), and can be used to compose models zero-shot.

Despite these advantages, training EBMs with contrastive divergence has been challenging due to training instabilities. Ensuring training stability required either combinations of spectral normalization and Langevin dynamics gradient clipping \citep{du2019implicit}, parameter tuning \citep{grathwohl2019your}, early stopping of MCMC chains \citep{nijkamp2019learning}, or avoiding the use of modern deep learning components, such as self-attention or layer normalization \citep{du2019implicit}. These requirements limit modeling power, prevent the compatibility with modern deep learning architectures, and prevent long-running training procedures required for scaling to larger datasets. With this work, we aim to maintain the simplicity and advantages of contrastive divergence training, while resolving stability issues and incorporating complementary deep learning advances.

\begin{figure}[t]
\includegraphics[width=1\linewidth]{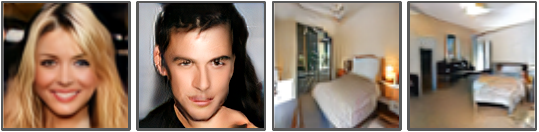}
\vspace{-15pt}
\caption{\small{(Left) 128x128 samples on unconditional CelebA-HQ. (Right)  128x128  samples on unconditional LSUN Bedroom. }}
\label{fig:teaser_img}
\vspace{-20pt}
\end{figure}

An often overlooked detail of contrastive divergence formulation is that changes to the energy function change the MCMC samples, which introduces an additional gradient term in the objective function (see \sect{sect:missing} for details). This term was claimed to be empirically negligible in the original formulation and is typically ignored \citep{Hinton2002Training,liu2017learning} or estimated via high-variance likelihood ratio approaches \citep{ruiz2019contrastive}. We show that this term can be efficiently estimated for continuous data via a combination of auto-differentiation and nearest-neighbor entropy estimators. We also empirically show that this term contributes significantly to the overall training gradient and has the effect of stabilizing training. It enables inclusion of self-attention blocks into network architectures, removes the need for capacity-limiting spectral normalization, and allows us to train the networks for longer periods. We do not introduce any new objectives or complexity - our procedure is simply a more complete form of the original formulation.

We further present techniques to improve mixing and mode exploration of MCMC transitions in contrastive divergence. We propose data augmentation as a useful tool to encourage mixing in MCMC by directly perturbing input images to related images. By incorporating data augmentation as semantically meaningful perturbations, we are able to greatly improve mixing and diversity of MCMC chains. We also leverage compositionality of EBMs to evaluate an image sample at multiple image resolutions when computing energies. Such evaluation and coarse and fine scales leads to samples with greater spatial coherence, but leaves MCMC generation process unchanged. We note that such hierarchy does not require specialized mechanisms such as progressive refinement \citep{Karras2017Progressive}

Our contributions are as follows: firstly, we show that a gradient term neglected in the popular contrastive divergence formulation is both tractable to estimate and is important in avoiding training instabilities that previously limited applicability and scalability of energy-based models. Secondly, we highlight how data augmentation and multi-scale processing can be used to improve model robustness and generation quality. Thirdly, we empirically evaluate stability of model architectures and show improved performance on a host of benchmarks and use cases, such as image generation, OOD detection, and compositional generation\footnote{Project page and code: https://energy-based-model.github.io/improved-contrastive-divergence/}.

\section{An Improved Contrastive Divergence Framework for Energy-Based Models}
Energy-Based Models (EBMs) represent the likelihood of a probability distribution $p_D(\vx)$ for $\vx \in \R^D$ as $p_{\theta}(\vx) = \frac{\exp(-E_{\theta}(\vx))}{Z(\theta)}$ where the function $E_{\theta}(\vx): \R^D \rightarrow \R$, is known as the \textit{energy function}, and $Z(\theta) = \int_{\vx} \exp{-E_{\theta}(\vx)}$ is known as the partition function. Thus, an EBM can be represented by an neural network that takes $\vx$ as input and outputs a scalar.

Training an EBM through maximum likelihood (ML) is not straightforward, as $Z(\theta)$ cannot be reliably computed, since this involves  integration over the entire input domain of $\vx$. However, the gradient of log-likelihood with respect to a data sample $\vx$ can be represented as 
\begin{equation}
    \label{eq:aml}
    \resizebox{0.85\hsize}{!}{$
    \frac{\partial \log p_{\theta} (\vx)}{\partial \theta} = -\left( \frac{\partial E_\theta(\vx)}{\partial \theta} - \E_{p_{\theta}(\vx')} \left [\frac{ \partial E_\theta(\vx')}{\partial \theta} \right] \right) $}. 
\end{equation}
Note that \eqn{eq:aml} is still not tractable, as it requires using Markov Chain Monte Carlo (MCMC) to draw samples from the model distribution $p_{\theta}(\vx)$, which often takes exponentially long to mix. As a practical approximation to the above objective, \citep{Hinton2002Training} proposes the contrastive divergence objective
\begin{equation}
    \kldiv{p_D(\vx)}{p_{\theta} (\vx)} - \kldiv{\Pi_\theta^t(p_D(\vx))}{p_{\theta}(\vx)}, 
    \label{eq:cd}
\end{equation}
where $\Pi_\theta$ represents a MCMC transition kernel for $p_{\theta}$, and $\Pi_\theta^t(p_D(\vx))$ represents $t$ sequential MCMC transitions starting from $p(\vx)$.  In this objective, if we can guarantee that%
\begin{equation}
    \kldiv{p_D(\vx)}{p_{\theta} (\vx)} \ge \kldiv{\Pi^t_\theta(p_D(\vx))}{p_{\theta}(\vx)},
    \label{eqn:ge}
\end{equation}%
then the objective guarantees that $p_{\theta}(\vx)$ converges to the data distribution $p_D(\vx)$, since the objective is only zero (at its fixed point) when $p_{\theta}(\vx) = p_D(\vx)$.

If $\Pi$ represents a MCMC transition kernel, this property is guaranteed \citep{lyu2011unifying}. Note that $\Pi$ \textit{does not} need to converge to the underlying probability distribution, and only a finite number of steps of MCMC sampling may be used. In fact, this objective may be utilized to maximize likelihood even if $\Pi$ is \textit{not} a MCMC transition kernel, but instead a model such as an amortized generator, as long as we ensure that \eqn{eqn:ge} holds. In the appendix, we show that our approach applies even when MCMC chains are not initialized from the data distribution. 

\begin{figure*}
\centering
\includegraphics[width=1\linewidth]{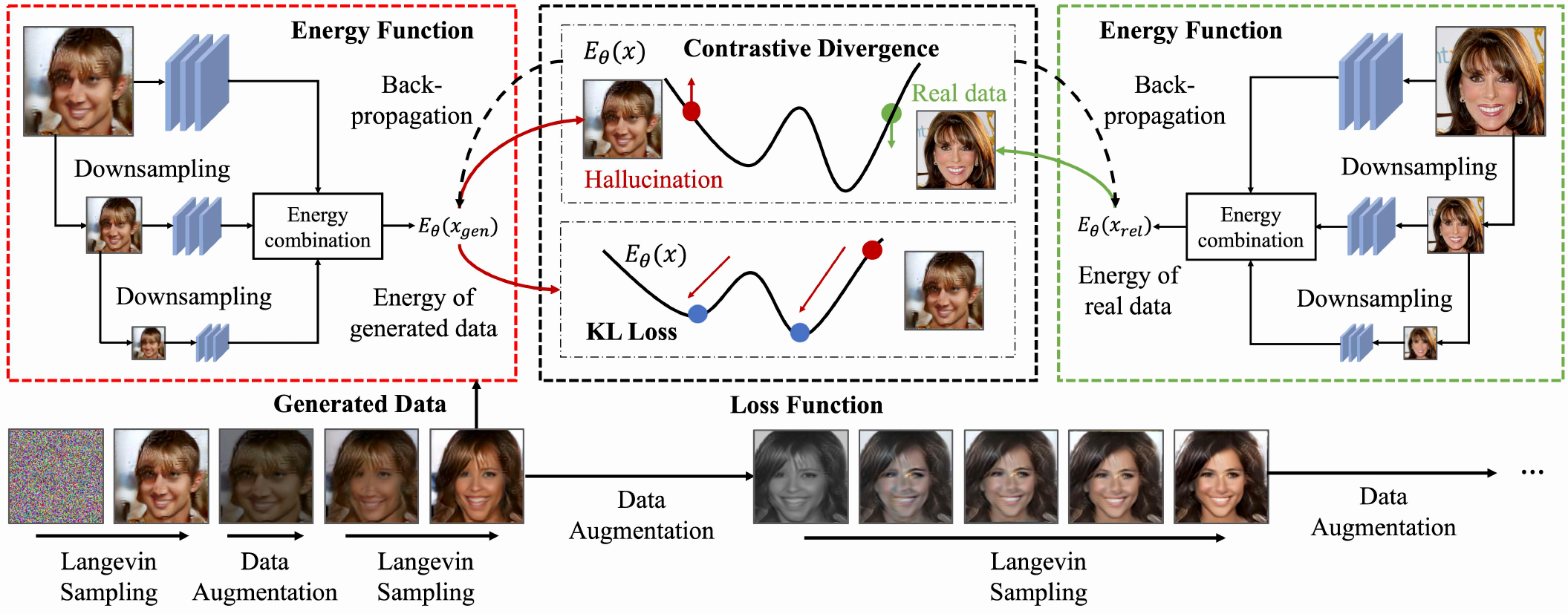}
\vspace{-15pt}
\caption{\small Illustration of our overall proposed framework for training EBMs. EBMs are trained with contrastive divergence, where the energy function decreases energy of real data samples (green dot) and increases the energy of hallucinations (red dot). EBMs are further trained with a KL loss which encourages generated hallucinations (shown as a solid red ball) to have low underlying energy and high diversity (shown as blue balls). Red/green arrows indicate forward computation while dashed arrows indicate gradient backpropogation.}
\label{fig:pipeline_img}
\vspace{-15pt}
\end{figure*}

\subsection{A Missing Term in Contrastive Divergence}
\label{sect:missing}
When taking the negative gradient of the contrastive divergence objective (\eqn{eq:cd}), we obtain the expression%
\begin{equation}
    \label{eq:kl_grad}
    \resizebox{0.8\hsize}{!}{$
    \begin{split}
    - ( \E_{p_D(\vx)} \left [\frac{\partial E_{\theta}(\vx)}{\partial \theta} \right] &- \E_{q_{\theta}(\vx')} [\frac{\partial E_{\theta}(\vx')}{\partial \theta}] \\
    &+ {\color{red} \frac{\partial q(\vx')}{\partial \theta} \frac{\partial \kldiv{q_{\theta}(\vx')}{p_{\theta}(\vx')}}{\partial q_{\theta}(\vx')} } )
    \end{split} $},
\end{equation}
where for brevity, we summarize $\Pi_\theta^t(p(\vx)) = q_{\theta}(\vx)$. The first two terms are identical to those of \eqn{eq:aml} and the third gradient term (which we refer to as the {\color{red}KL divergence term}) corresponds to minimizing the divergence between $q_{\theta}(\vx)$ and $p_{\theta}(\vx)$. In practice, past contrastive divergence approaches have ignored the third gradient term, which was difficult to estimate and claimed to be empirically negligible \citep{hinton1999products} (which in \fig{fig:kl_cd} we show to be non-negligible), leading 
to the incorrect optimization of \eqn{eq:cd}. To correctly optimize \eqn{eq:cd}, we construct a new joint loss expression $\mathcal{L}_{\text{Full}}$,  consisting of traditional contrastive loss $\mathcal{L}_{\text{CD}}$ and a new loss expression $\mathcal{L}_{\text{KL}}$, to accurately exhibit all three gradient terms. Specifically, we have
$\mathcal{L}_{\text{Full}} = \mathcal{L}_{\text{CD}} + \mathcal{L}_{\text{KL}}$ where $\mathcal{L}_{\text{CD}}$ is
\begin{equation}
    \mathcal{L}_{\text{CD}} =  \E_{p_D(\vx)} [E_{\theta}(\vx)] - \E_{\text{stop\_grad}(q_{\theta}(\vx'))} [E_{\theta}(\vx')],
    \label{eq:cd_loss}
\end{equation}
and the ignored KL divergence term corresponding to the following KL loss:
\begin{equation}
    \resizebox{0.87\hsize}{!}{$
    \mathcal{L}_{\text{KL}} =  \E_{q_{\theta}(\vx)} [E_{\text{stop\_grad}(\theta)}(\vx)] + \E_{q_{\theta}(\vx)}[\log (q_{\theta}(\vx))] $}.
    \label{eq:kl_loss}
\end{equation}
Despite being difficult to estimate, we show that $\mathcal{L}_{\text{KL}}$ is a useful tool for both speeding up and stabilizing training of EBMs.  We provide derivations showing the equivalence of gradients of $\mathcal{L}_{\text{Full}}$ and that of \eqn{eq:cd} in the appendix, where stop gradient operators are necessary to ensure correct gradients. \fig{fig:pipeline_img} illustrates  the overall effects of both losses. \eqn{eq:cd_loss} encourage the energy function to assign low energy to real samples and high energy for generated samples. However, only optimizing \eqn{eq:cd_loss} often leads to an adversarial mode where the energy function learns to simply generate an energy landscape that makes sampling difficult. The KL divergence term counteracts this effect and encourages sampling to closely approximate the underlying distribution $p_\theta(\vx)$, by encouraging samples to be both low energy under the energy function as well as diverse. Empirically, we find that including for KL term significantly improves both the stability, generation quality, and robustness to different model architectures (\fig{fig:kl_loss}). Next, we will discuss our approach towards estimating this KL divergence. 

\subsection{Estimating the missing gradient term} 

Estimating $\mathcal{L}_{\text{KL}}$ can further be decomposed into two separate objectives, minimizing the energy of samples from $q_{\theta}(\vx)$, which we refer to as $\mathcal{L}_{\text{opt}}$ (\eqn{eq:opt_loss}) and maximizing the entropy of samples from $q_{\theta}(\vx)$ which we refer to as $\mathcal{L}_{\text{ent}}$ (\eqn{eq:ent_loss}). 

\textbf{Minimizing Sampler Energy.} To minimize the energy of samples from $q_{\theta}(\vx)$ we can directly differentiate through both the energy function and MCMC sampling. We follow recent work in EBMs and utilize Langevin dynamics \citep{du2019implicit, nijkamp2019learning, grathwohl2019your} for our MCMC transition kernel, and note that each step of Langevin sampling is fully differentiable with respect to underlying energy function parameters. Precisely, gradient of  $\mathcal{L}_{\text{opt}}$,  $\frac{ \partial \mathcal{L}_{\text{opt}}}{ \partial \theta}$ becomes
\begin{equation}
    \label{eq:opt_loss}
    \resizebox{0.9\hsize}{!}{$
     \E_{q_{\theta}(\vx_0', \ldots, \vx_t')} \left [\frac{\partial E_{\text{stop\_grad}(\theta)} (\vx_{t-1}' - \nabla_{\vx_{t-1}'} E_{\theta} (\vx_{t-1}') + \omega) }{\partial \theta} \right], 
    $}
\end{equation}

where $\omega \sim \mathcal{N}(0,\lambda)$ and $\vx_i'$ represents the $\text{i}^{\text{th}}$ step of Langevin sampling. To reduce the memory overhead of this differentiation procedure, we only differentiate through the last step of Langevin sampling as also done in \citep{pmlr-v119-vahdat20a}. In the appendix we show that this leads to the same effect as differentiation through Langevin sampling.

\textbf{Entropy Estimation.} To maximize the entropy of samples from $q_{\theta}(\vx)$, we use a non-parametric nearest neighbor entropy estimator \citep{beirlant1997nonparametric}, which is shown to be mean square consistent \citep{kozachenko1987sample} with root-n convergence rate \citep{tsybakov1996root}.  The entropy $H$ of a distribution $p(\vx)$ can be estimated through a set $X = {x_1, x_2, \ldots, x_n}$ of $n$ different points sampled from $p(\vx)$ as $H(p_{\theta}(\vx)) = \frac{1}{n} \sum_{i=1}^n \ln (n \cdot \operatorname{NN}(x_i, X)) + O(1)$ where the function $\operatorname{NN}(x_i, X)$ denotes the nearest neighbor distance of $x_i$ to any other data point in $X$. Based off the above entropy estimator, we write $\mathcal{L}_{\text{ent}}$ as the entropy loss:

\begin{equation}
    \mathcal{L}_{\text{ent}} = \E_{q(\vx)}[-\log ( \operatorname{NN}(\vx, B))]
      \label{eq:ent_loss}
\end{equation}
where we measure the nearest neighbor with respect to a set $B$ of 100 past samples from MCMC chains. We utilize L2 distance as the metric for computing nearest neighbors. This type of nearest entropy estimator is known to scale poorly to high dimensions (requiring an exponential number of samples to yield an accurate entropy estimate). However, in our setting, we do not need an accurate estimate of entropy. Instead, our computation of entropy is utilized as a fast regularizer to prevent sampling from collapsing.

\subsection{Data Augmentation Transitions}
Langevin sampling, our MCMC transition kernel, is prone to falling into local probability modes \citep{neal2011mcmc}. In the image domain, this manifests with sampling chains always converging to a fixed image \citep{du2019implicit}.  A core difficulty is that distances between two qualitatively similar images can be significantly far away from each other in the input domain, on which sampling is applied. While $\mathcal{L}_{\text{KL}}$ serves as a regularizer to prevent sampling collapse in Langevin dynamics,   Langevin dynamics alone is not enough to encourage large jumps in a finite number of steps. It is further beneficial to have an individual sampling chain have the ability to mix between probability modes.

 \begin{algorithm}
\small
\begin{algorithmic}
    \STATE \textbf{Input:} data dist $p_D(\vx)$, step size $\lambda$, number of steps $K$, data augmentation $D(\cdot)$, stop gradient operator $\Omega(\cdot)$, EBM $E_\theta(\cdot)$
    \STATE $\mathcal{B} \gets \varnothing$
    \WHILE{not converged}
    \STATE $\vx^+_i \sim p_D$
    \STATE $\tilde{\vx}^0_i \sim \mathcal{B}$ with 99\% probability and $\mathcal{U}$ otherwise
    \STATE $X \sim \mathcal{B}$ for nearest neighbor entropy calculation
    \vspace{2mm}
    \STATE \emph{$\triangleright$ Apply data augmentation to sample:}
    \STATE $\tilde{\vx}^0_i = D(\tilde{\vx}^0_i)$ 
    
    \vspace{2mm}
    \STATE \emph{$\triangleright$ Generate sample using Langevin dynamics:}
    \FOR{sample step $k = 1$ to $K$}
    \STATE $\tilde{\vx}^{k-1}_i = \Omega(\tilde{\vx}^{k-1}_i) $ 
    \STATE $\tilde{\vx}^k \gets \tilde{\vx}^{k-1} -  \nabla_\vx E_\theta (\tilde{\vx}^{k-1}) + \omega, \;\; \omega \sim \mathcal{N}(0,\sigma)$
    \ENDFOR 
    \vspace{2mm}
    \STATE \emph{$\triangleright$ Generate two variants of $\vx^-$ with and without gradient propagation:}
    \STATE $\vx^-_i = \Omega(\tilde{\vx}^k_i) $ 
    \STATE $\hat{\vx}^-_i = \tilde{\vx}^k_i $ 
    \vspace{2mm}
    \STATE \emph{$\triangleright$ Optimize objective $\mathcal{L}_\text{CD} + \mathcal{L}_\text{KL}$ wrt $\theta$:}
    \STATE $\mathcal{L}_{\text{CD}} = \frac{1}{N} \sum_i  ( E_\theta(\vx^+_i) - E_\theta(\vx^-_i)$ 
    \STATE $\mathcal{L}_{\text{KL}} = E_{\Omega(\theta)}(\hat{\vx}^-_i) - \log(NN(\hat{\vx}^-_i, X)$ 
    
    \vspace{2mm}
    \STATE \emph{$\triangleright$ Optimize objective $\mathcal{L}_\text{CD} + \mathcal{L}_\text{KL}$ wrt $\theta$:}
    \STATE $\Delta \theta \gets \nabla_\theta (\mathcal{L}_{\text{CD}} + \mathcal{L}_{\text{KL}})$
    \STATE Update $\theta$ based on $\Delta \theta$ using Adam optimizer  
    \vspace{2mm}

    \STATE \emph{$\triangleright$ Update replay buffer $\mathcal{B}$}
    \STATE $\mathcal{B} \gets \mathcal{B} \cup \tilde{\vx}_i^-$
    \ENDWHILE
  \end{algorithmic}
 \caption{EBM training algorithm}
 \label{alg:energy}
 \end{algorithm} 
\begin{figure}

\centering
\includegraphics[width=0.85\linewidth]{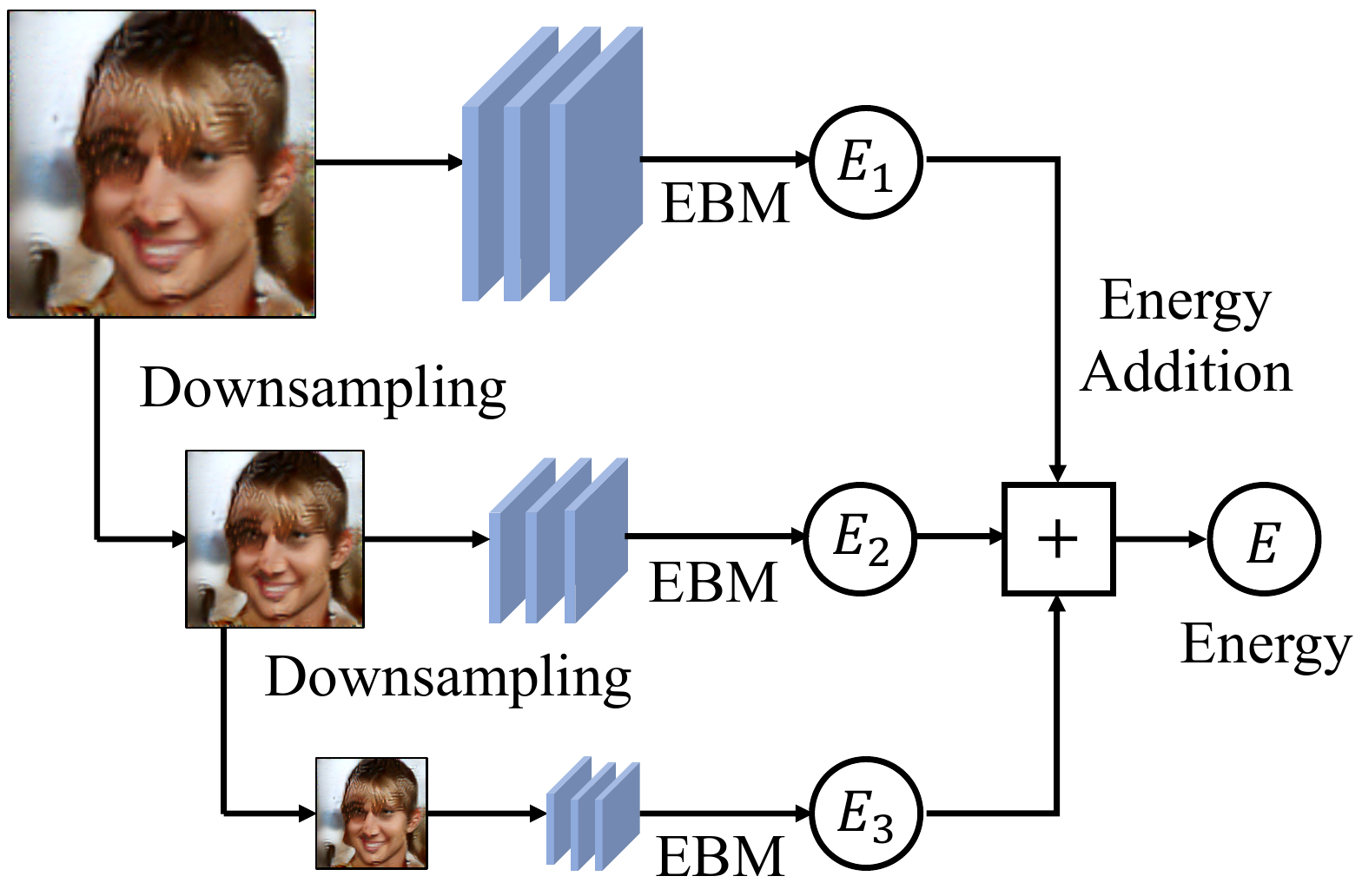}
\vspace{-5pt}
\caption{\small Illustration of our multi-scale EBM architecture. Our energy function over an image is defined compositionally as the sum of energy functions on different resolutions of an image.}
\label{fig:multi-scale}
\vspace{-20pt}
\end{figure}

To encourage greater exploration between similar inputs in our model, we propose to augment chains of MCMC sampling with periodic data augmentation transitions that encourages movement between ``similar'' inputs. In particular, we utilize a combination of color, horizontal flip, rescaling, and Gaussian blur augmentations. Such combinations of augmentation has recently seen success applied in unsupervised learning \citep{chen2020simple}. Specifically, during training time,  we initialize MCMC sampling from a data augmentation applied to an input sampled from the buffer of past samples. At test time, during the generation, we apply a random augmentation to the input after every 20 steps of Langevin sampling. We illustrate this process in the bottom of \fig{fig:pipeline_img}. Data augmentation transitions are always taken.

\subsection{Compositional Multi-scale Generation}

To encourage energy functions to focus on features in both low and high resolutions, we define our energy function as the composition (sum) of a set of energy functions operating on different scales of an image, illustrated in \fig{fig:multi-scale}. Since the downsampling operation is fully differentiable, Langevin based sampling can be directly applied to the energy function. In our experiments, we utilize full, half, and quarter resolution image as input and show that this improves the generation performance.

\begin{algorithm}
\small
\begin{algorithmic}
    \STATE \textbf{Input:} number of data augmentation applications $N$, step size $\lambda$, number of steps $K$, data augmentation $D(\cdot)$, EBM $E_\theta(\cdot)$
    \STATE $\tilde{\vx}^0 \sim \mathcal{U}$
    \vspace{2mm}
    
    \STATE \emph{$\triangleright$ Generate samples through $N$ iterative steps of data augmentation/Langevin dynamics:}
    \FOR{sample step $n = 1$ to $N$}
    \STATE \emph{$\triangleright$ Apply data augmentation to samples:}
    \STATE $\tilde{\vx}^0 = D(\tilde{\vx}^0_i)$
    \vspace{2mm}
    \STATE \emph{$\triangleright$ Run $K$ steps of  Langevin dynamics:}
    \FOR{sample step $k = 1$ to $K$}
    \STATE $\tilde{\vx}^k \gets \tilde{\vx}^{k-1} -  \nabla_\vx E_\theta (\tilde{\vx}^{k-1}) + \omega, \;\; \omega \sim \mathcal{N}(0,\sigma)$
    \ENDFOR 
    \vspace{2mm}
    \STATE \emph{$\triangleright$ Iteratively refine samples:}
    \STATE $\tilde{\vx}^0 = \tilde{\vx}^k$
    \ENDFOR 
    \vspace{2mm}
    \STATE \emph{$\triangleright$  Final output:}
    \STATE $\vx = \tilde{\vx}^0$
  \end{algorithmic}
 \caption{EBM sampling algorithm}
 \label{alg:sample}
 \end{algorithm}

\subsection{Training Algorithm and Sampling}
We provide an overview of our overall proposed training algorithm in Algorithm \ref{alg:energy}. Our overall approach is  similar to the algorithm presented in \citep{du2019implicit}, with two notable differences. First, we apply data augmentation to samples drawn for the replay buffer. Second, we propagate gradients through sampling to  efficiently compute $\mathcal{L}_{\text{KL}}$.  We further present the sample generation algorithm for a trained EBMs in Algorithm \ref{alg:sample}. We iteratively apply $N$ steps of data augmentation and Langevin sampling to mimic the replay buffer utilized during training.

\vspace{-10pt}
\section{Experiments}
\label{sec:experiments}
We perform empirical experiments to validate the following set of questions: (1) What are the effects of each proposed component towards training EBMs? (2) Are our trained EBMs able to perform well on downstream applications of EBMs, such as image generation, out-of-distribution detection, and concept compositionality?

\subsection{Experimental Setup}
We investigate the efficacy of our proposed approach. Models are trained using the Adam Optimizer \citep{Kingma2015Adam}, on a single 32GB Volta GPU for CIFAR-10 for 1 day, and for 3 days on 8 32GB Volta GPUs for CelebaHQ, LSUN and ImageNet 32x32 datasets. We provide detailed training configuration details in the appendix. 

Our improvements are largely built on top of the EBMs training framework proposed in \citep{du2019implicit}. We use a buffer size of 10000, with a resampling rate of 0.1\%. Our approach is significantly more stable than IGEBM, allowing us to remove aspects of regularization in \citep{du2019implicit}. We remove the clipping of gradients in Langevin sampling as well as spectral normalization on the weights of the network. In addition, we add self-attention blocks and layer normalization blocks in residual networks of our trained models. In multi-scale architectures, we utilize 3 different resolutions of an image, the original image resolution, half the image resolution and a quarter the image resolution. We report detailed architectures in the appendix. When evaluating models, we utilize the EMA model with EMA weight of 0.9999. 

\begin{table}
    \centering
    \caption{\small Table of Inception and FID scores for generations of CIFAR-10, CelebA-HQ and ImageNet32x32 images. All others numbers  are taken directly from corresponding papers. On CIFAR-10, our approach outperforms past EBM approaches and achieves performance close to SNGAN. On CelebA-HQ, our approach achieves performance close to that of SSGAN.  On ImageNet 32x32, our approach achieves similar performance to the PixelIQN (large) model with around one tenth the parameters.}
    \resizebox{\columnwidth}{!}{
    \begin{tabular}{lcc}
    \toprule
    Model & Inception & FID\\
    \midrule
    \midrule
    \textbf{CIFAR-10 Unconditional} & & \\
    \midrule
    PixelCNN \citep{VanOord2016Pixel} & 4.60 & 65.9  \\
    Multigrid EBM \citep{gao2018learning} & 6.56 &  40.1\\
    IGEBM (Ensemble) \citep{du2019implicit} & 6.78 & 38.2 \\
    Short-Run EBM \citep{nijkamp2019learning} & 6.21 &  44.5\\
    DCGAN \citep{Radford2016Unsupervised} & 6.40  & 37.1 \\
    WGAN + GP \citep{Gulrajani2017Improved} &  6.50 &  36.4\\
    NCSN \citep{song2019generative} & 8.87 & 25.3 \\
    Ours & 7.85 & 25.1 \\
    SNGAN \citep{miyato2018spectral} & 8.22 & 21.7 \\
    SSGAN \citep{chen2019self} & - & 19.7 \\
    \midrule
    \textbf{CelebA-HQ 128x128 Unconditional} & & \\
    \midrule
    Ours  & - & 28.78 \\
    SSGAN \citep{chen2019self} & - & 24.36 \\
    \midrule
    \textbf{ImageNet 32x32 Unconditional} & & \\
    PixelCNN \citep{Oord2016Conditional} & 7.16 & 40.51 \\
    PixelIQN (small) \citep{ostrovski2018autoregressive} & 7.29 & 37.62 \\
    PixelIQN (large) \citep{ostrovski2018autoregressive} & 8.69 & 26.56 \\
    IGEBM \citep{du2019implicit} & 5.85 & 62.23 \\
    Ours & 8.73 & 32.48  \\
    \bottomrule
    \end{tabular}
    }
    \label{tbl:main_quant}
\end{table}

\begin{figure}
    \centering
    \includegraphics[width=1.0\linewidth]{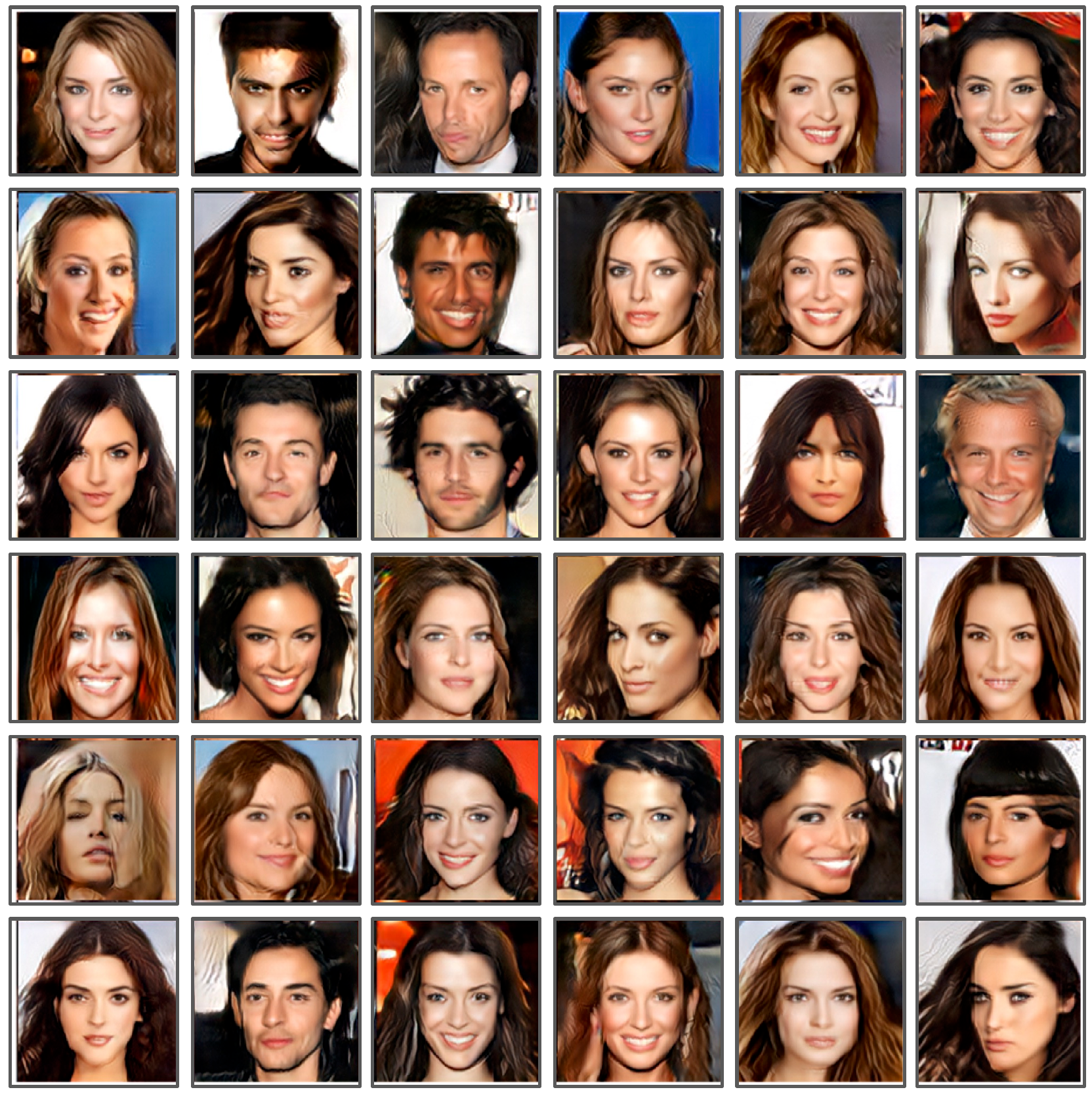}
    \vspace{-10pt}
    \caption{\small 
    Randomly selected unconditional 128x128 CelebA-HQ images generated from our trained EBM model. Samples are relatively diverse with limited artifacts.
    } 
    \label{fig:celeba}
    \vspace{-10pt}
\end{figure}

\begin{figure*}
\centering
\includegraphics[width=1\linewidth]{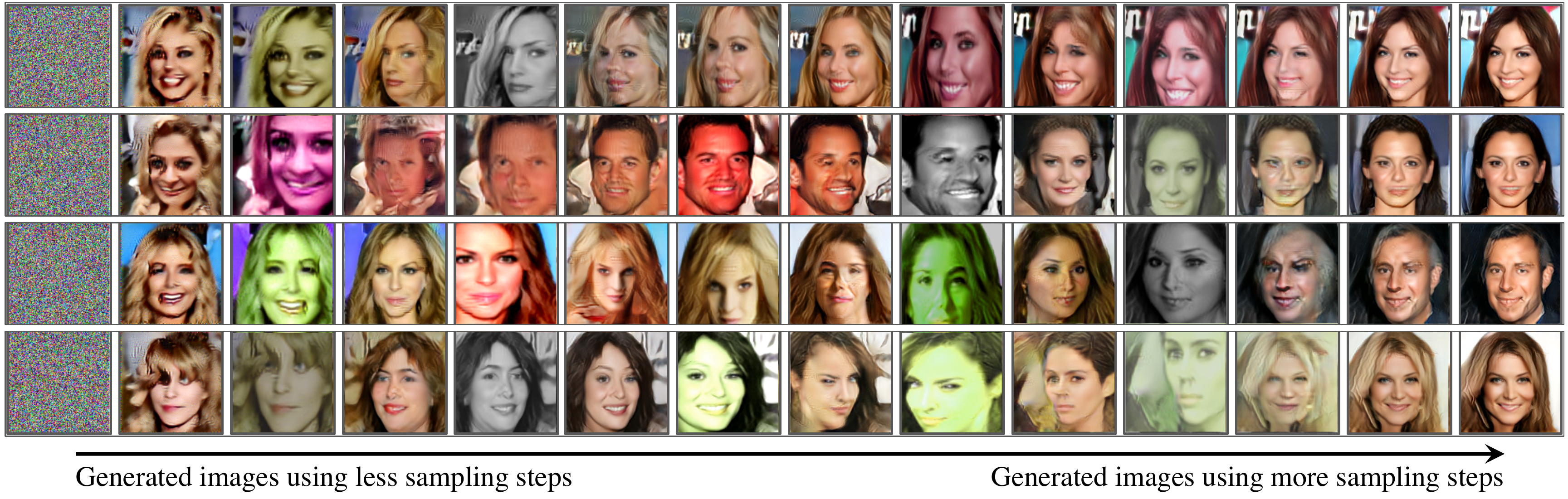}
\vspace{-15pt}
\caption{\small Visualization of Langevin dynamics sampling chains on an EBM trained on CelebA-HQ 128x128. Samples travel between different modes of images. Each consecutive images represents 30 steps of sampling, with data augmentation transitions every 60 steps.}
\label{fig:sample_init}
\vspace{-10pt}
\end{figure*}
\subsection{Image Generation}

\begin{figure}[t]
\centering
\includegraphics[width=0.95\linewidth]{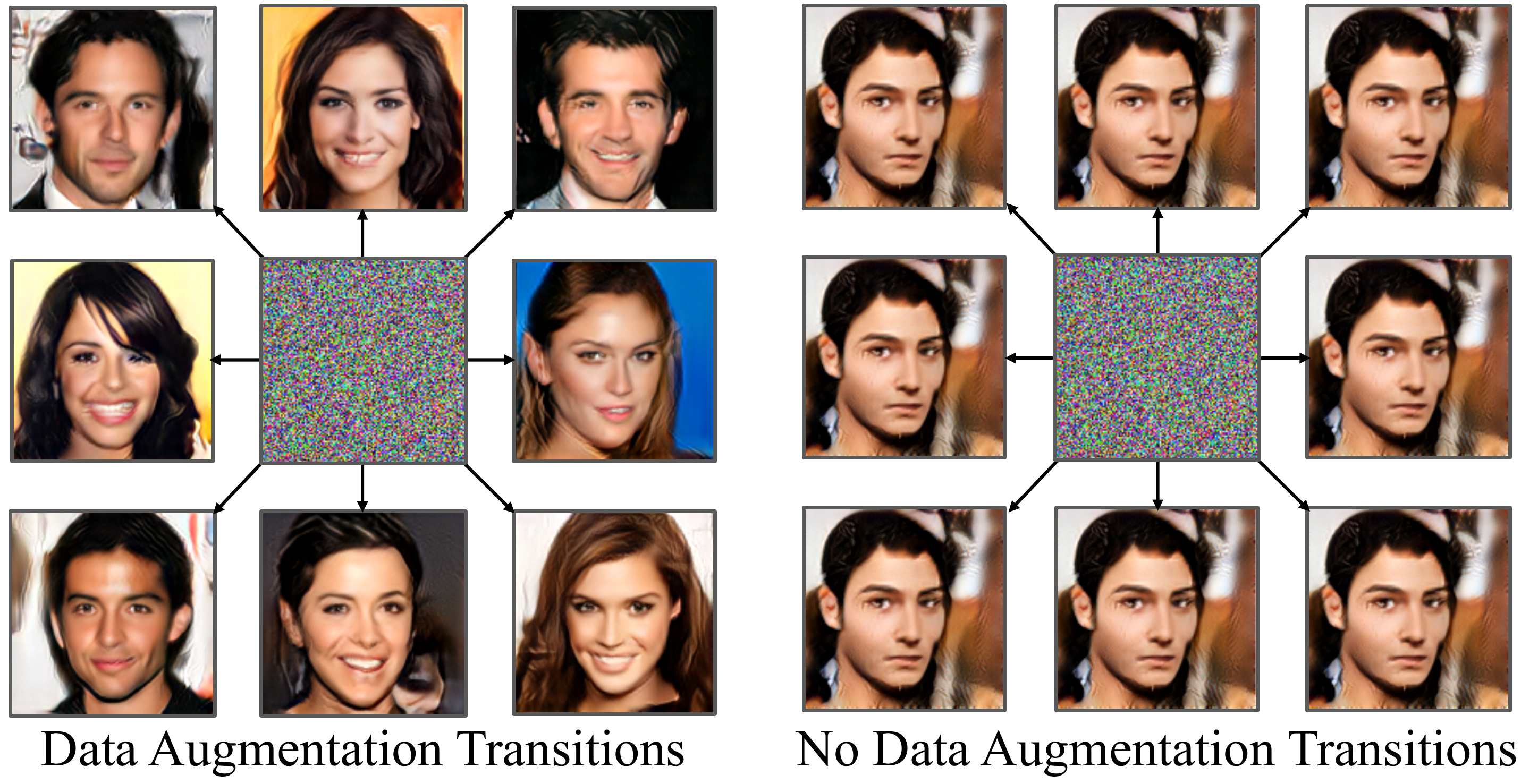}

\caption{\small Output samples after running Langevin dynamics from a fixed initial sample (center of square), with or without intermittent data-augmentation transitions. Without data-augmentation transitions, all samples converge to same image, while data augmentations enables chains to seperate.}
\label{fig:diverse_sample}
\vspace{-10pt}
\end{figure}

\begin{figure*}
    \centering
    \includegraphics[width=1\linewidth]{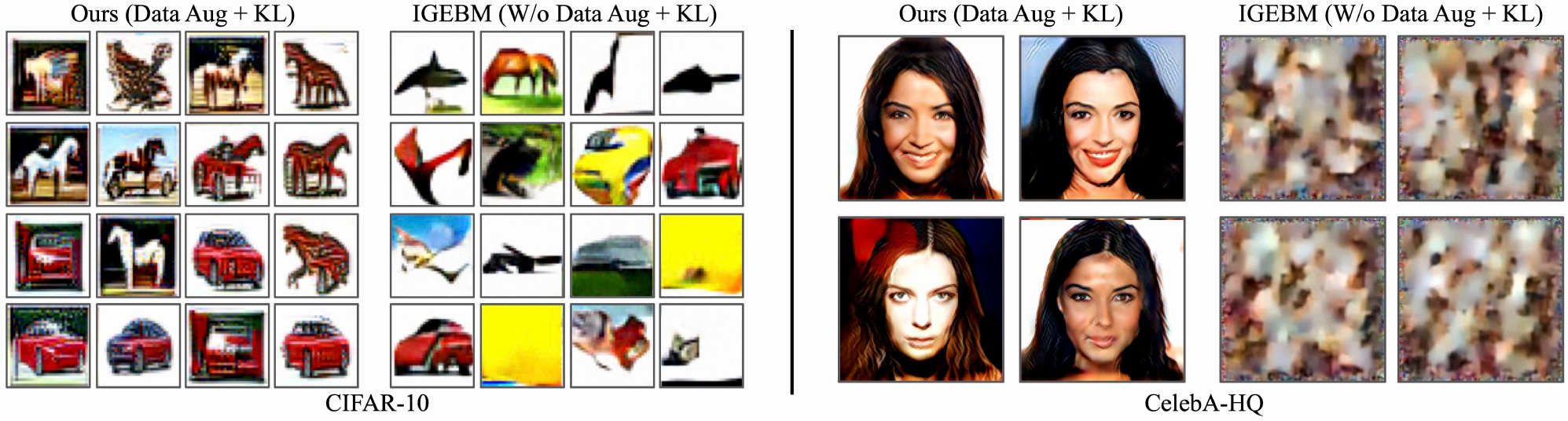}
    \vspace{-15pt}
    \caption{\small Illustration of very low temperature samples from our model with KL loss and data augmentation compared to those from IGEBM on CIFAR-10 (left) and CelebA-HQ (right). After a large number of sampling steps,  IGEBM converges to stranges hues in CIFAR-10 and random textures on CelebA-HQ. In contrast, due to better mode exploration, adding both improvements maintains naturalistic image modes on both CIFAR-10 and CelebA-HQ.}
    \label{fig:mode_convergence}
    \vspace{-10pt}
\end{figure*}

\begin{figure}[t]
    \centering
    \includegraphics[width=1\linewidth]{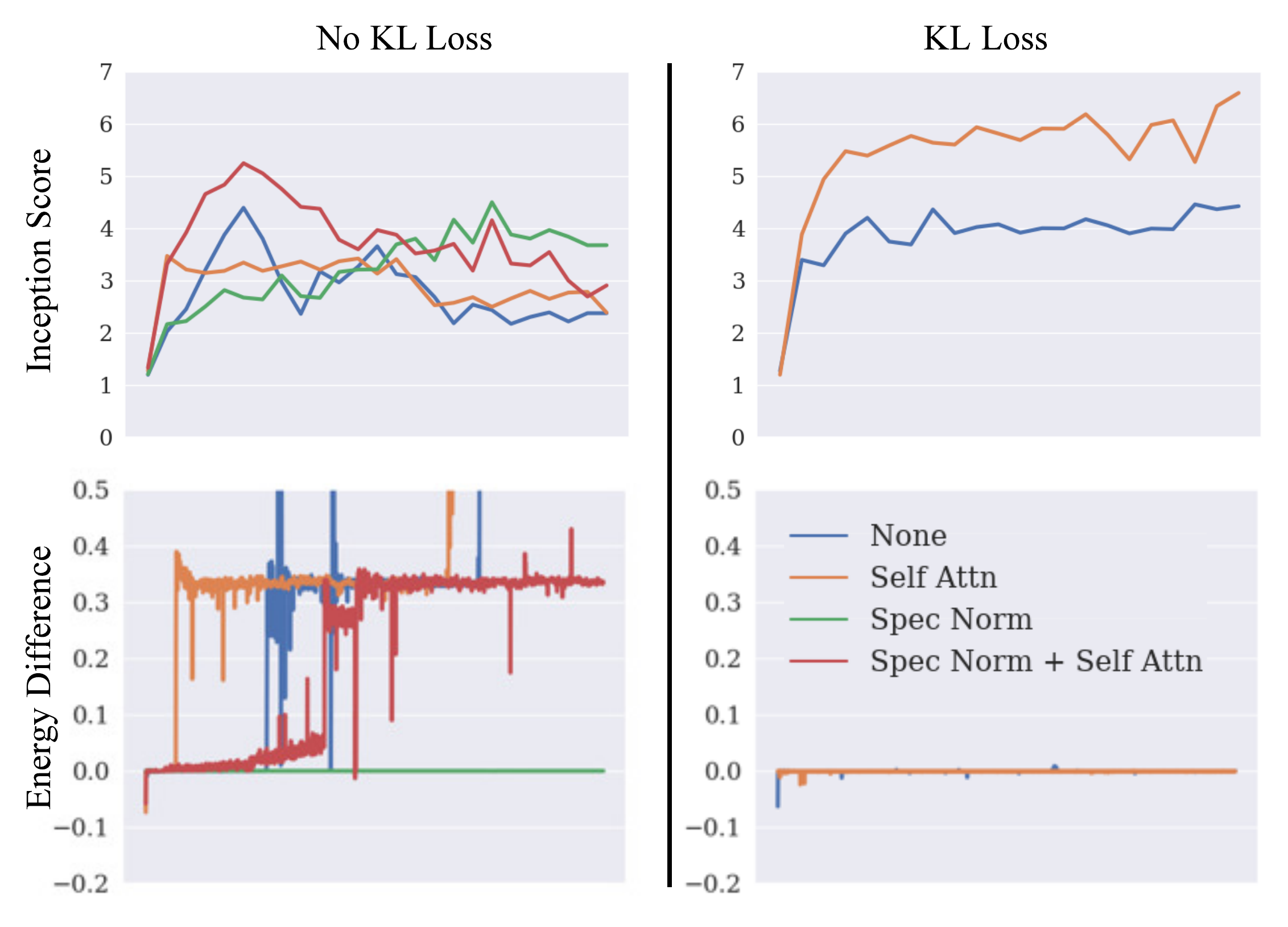}
    \vspace{-15pt}
    \caption{\small The KL loss significantly improves the stability of EBM training. Stable EBM training occurs when the energy difference (illustrated in bottom row) is roughly zero. We find that without using the KL loss term (left column), EBM training quickly diverges (bottom left). Spectral normalization prevents divergence of energies, but cannot be combined with self attention without destabilizing training. KL loss (right column) maintains an energy differences to 0 (bottom right), even with the addition of self attention. Inception scores rapidly rises with KL loss (top right), but fall without KL loss (top left) due to destabilized training. Spectral norm prevents the Inception score from falling, but the score also does not increase much due to constraints in the architecture.}
    \label{fig:kl_loss}
    \vspace{-10pt}
\end{figure}

We evaluate our approach on  CIFAR-10, Imagenet 32x32 \citep{Deng2009Imagenet}, and CelebA-HQ \citep{Karras2017Progressive} datasets. Additional quantitative comparisons, results, and ablations can be found in the appendix of the paper.

\textbf{Image Quality.} We evaluate our approach on unconditional generation in \tbl{tbl:main_quant}. We utilize Inception \citep{Salimans2016Improved} and FID \citep{heusel2017gans} implementations from \citep{du2019implicit} to evaluate samples. On CIFAR-10, we find that our approach outperforms many past EBM approaches in both FID and Inception scores using a similar number of parameters. We find that our performance is slightly worse than that of SNGAN and SSGAN.  On CelebA-HQ, we find that our approach outperforms our reimplementation of SNGAN using default ImageNet hyperparameters, and is close to the reported numbers of SSGAN. Finally, on Imagenet32x32, we find that our approach outperforms previous EBM models and achieves performance comparable to that of the large PixelIQN model in terms of FID and Inception score. We note, however, that our model is significantly smaller than the PixelIQN model and has one tenth the number of parameters.  We present example qualitative images from CelebA-HQ in \fig{fig:celeba} and present qualitative images on other datasets in the appendix of the paper.
While our overall generative performance are not the best reported, we emphasize that it improves existing generative performance of EBMs, which have unique benefits such as compositionality (\sect{sect:compositionality}).

\vspace{-3pt}
\textbf{Effect of Data Augmentation.} We evaluate the effect of data augmentation on sampling in EBMs. In \fig{fig:sample_init} we show that by combining Langevin sampling with data augmentation transitions, we are able to enable chains to mix across different images, whereas prior works have shown Langevin converging to fixed images. In \fig{fig:diverse_sample} we show that given a fixed random noise initialization, data augmentation transitions enable to reach a diverse number of different samples, while  sampling without data augmentation transitions leads all chains to converge to the same face.

\vspace{-3pt}
\textbf{Mode Convergence.} We further investigate high likelihood modes of our model. In \fig{fig:mode_convergence}, we compare very low energy samples (obtained after running gradient descent 1000 steps on an energy function) for both our model with data augmentation and KL loss and the IGEBM model. Due to improved mode exploration, we find that low temperature samples under our model with data augmentation/KL loss reflect typical high likelihood "modes" in the training dataset, while our baseline models converges to odd shapes,  also noted in \citep{nijkamp2019anatomy}.

\vspace{-3pt}
\textbf{Stability/KL Loss.} EBMs are difficult to train and are sensitive to both the exact architecture and to various hyper-parameters. We found that the addition of a KL term into our training objective significantly improved the stability of training, by encouraging the sampling distribution to match the model distribution. In \fig{fig:kl_loss}, we measure training stability by measuring the energy differences between real and generated images. Stable training occurs when the energy difference is close to zero. Without $\mathcal{L}_{\text{KL}}$, we found that training our model with or without self-attention were both unstable, with differences spiking. Adding spectral normalization stabilizes training, but the addition of self-attention once again destabilizes training. In contrast with $\mathcal{L}_{\text{KL}}$, the addition of self-attention is also stable. We further compare Inception scores in \fig{fig:kl_loss} over training and find that while spectral normalization stabilizes training, it does at the expense of decreased improvement of Inception score. 

The addition of the KL term itself is not too expensive, simply requiring an additional nearest neighbor computation during training, a relatively insignificant cost compared to the number of negative sampling steps used during training. With a intermediate number of negative sampling steps (60 steps) during training, adding the KL term incurs a roughly 20\% computational cost. This difference is further decreased with a larger number of sampling steps. 

\vspace{-3pt}
\textbf{Ablations.} We ablate each portion of our proposed approach in \tbl{tbl:ablation}. We find that each our proposed components have significant gains in generation performance. In particular, we find a large gain in overall generative performance when adding the KL loss. This is in part due to a large boost in training stability (\fig{fig:kl_loss}), enabling significantly longer training times with both multiscale sampling and data augmentation. 
\begin{table}
    \centering
    \caption{\small Ablations of each proposed component on CIFAR-10 generation as well as corresponding stability of training. KL Loss significantly stabilizes EBM training, and enables larges boosts in generation performance via longer training. }
    \vspace{-5pt}
    \scalebox{0.7}{
    \begin{tabular}{ccccccc}
        \toprule
        KL Loss  & KL Loss & Data &  Multiscale & Inception  & FID & Stability\\
         $\mathcal{L}_{\text{opt}}$ & $\mathcal{L}_{\text{ent}}$ & Aug & Sampling & Score & & \\
        \midrule
        
        No & No & No & No & 3.57  & 169.74  & No \\
        No & No & Yes & No & 5.13 & 133.84  & No \\
        No & No & Yes & Yes & 6.14  & 53.78  & No \\
        Yes & No & Yes & Yes & 6.79 & 32.67  & Yes \\
        Yes & Yes & Yes & Yes & 7.85 & 25.08  & Yes \\
        \bottomrule
    \end{tabular}
    }
    \label{tbl:ablation}
    \vspace{-10pt}
\end{table}

\begin{figure}
\centering
\includegraphics[width=0.8\linewidth]{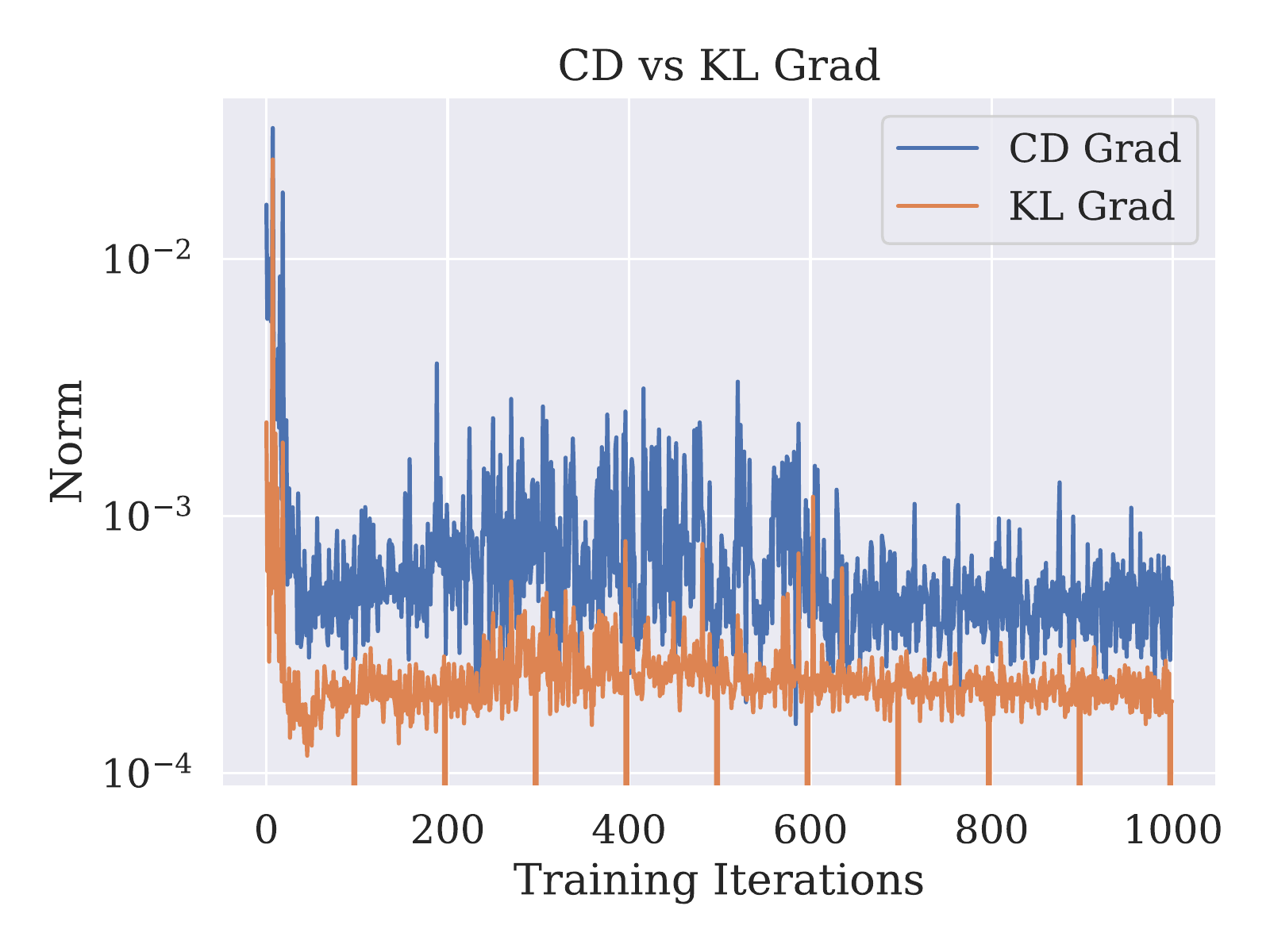}
\vspace{-13pt}
\caption{\small Plots of the gradient magnitude of $\gL_{\text{KL}}$ and  $\gL_{\text{CD}}$ across training iterations. Influences and relative magnitude of both loss terms stays constant through training.}
\label{fig:kl_cd}
\vspace{-20pt}
\end{figure}

\textbf{KL Gradient.} We plot the overall gradient magnitudes of $\mathcal{L}_{\text{CD}}$ and $\mathcal{L}_{\text{KL}}$ when training an EBM on CIFAR-10 in \fig{fig:kl_cd}. We find that relative magnitude of gradients of both training objectives remains constant across training, and that the gradient of the KL objective is non-negligible.

\begin{table}[t]
    \centering
    \begin{center}
    \caption{\small Table of AUROC values in out-of-distribution detection on unconditional models trained on CIFAR-10 using $\log(p_\theta(x))$. Our approach performs the best out of all methods. *JEM is not directly comparable as it uses supervised labels.} 
    \resizebox{\columnwidth}{!}{
    \begin{tabular}{l|cccccc}
        \toprule
        Model & PixelCNN++ & Glow & IGEBM  & JEM* & VERA &  Ours \\
        \midrule
        SVHN & 0.32 & 0.24 & 0.63 & 0.67 & 0.83 & \textbf{0.91} \\
        Textures & 0.33 & 0.27 & 0.48 & 0.60 & -  &\textbf{0.88} \\
        CIFAR10 Interp & 0.71 & 0.51 & 0.70 & 0.65 & \textbf{0.86} & 0.65\\
        CIFAR100 & 0.63 & 0.55 & 0.50 & 0.67 & 0.73 &  \textbf{0.83} \\
        Average & 0.50 & 0.39 & 0.57 & 0.65  & - &\textbf{0.82} \\
        \bottomrule
    \end{tabular}
    }
    \label{tbl:robustness}
    \end{center}
\end{table}

\vspace{-3pt}
\subsection{Out of Distribution Robustness}
\vspace{-3pt}
Energy-Based Models (EBMs) have also been shown to exhibit robustness to both out-of-distribution and adversarial samples \citep{du2019implicit, grathwohl2019your, grathwohl2020vera}.
We evaluate out-of-distribution detection of our trained energy function through log-likelihood using the AUROC evaluation metrics proposed in \citet{hendrycks2016baseline}. 
We similarly evaluate out-of-distribution detection of an unconditional CIFAR-10 model.

\textbf{Results.} 
We present out-of-distribution results in \tbl{tbl:robustness}, comparing with both likelihood models and EBMs using log-likelihood to detect outliers. We find that our approach significantly outperforms other baselines, with the exception of CIFAR-10 interpolations. We note the JEM \citep{grathwohl2019your} further requires supervised labels to train the energy function, which has to shown to improve out-of-distribution performance. We posit that by more efficiently exploring modes of the energy distribution at training time, we are able to reduce the spurious modes of the energy function and thus improve out-of-distribution performance.

\begin{table}[t]
    \centering
    \begin{center}
    \caption{\small Table of compositional generation accuracy across different models trained on the CelebA-HQ dataset. Generation accuracy measured through attribute predictions using a Resnet-18 classifier is trained to regress young, female, smiling, and wavy hair attributes in CelebA-HQ. Our approach achieves best performance.} 
    \resizebox{\columnwidth}{!}{
    \begin{tabular}{l|cccc}
        \toprule
        Model & Young & Female & Smiling & Wavy  \\
        \midrule
        JVAE (Young) & 0.543 & - & - & - \\
        JVAE (Young  $\&$ Female) & 0.440 & 0.554 & - & - \\
        JVAE (Young  $\&$ Female $\&$ Smiling) & 0.488 & 0.520 & 0.526   \\
        JVAE  (Young  $\&$ Female $\&$ Smiling $\&$ Wavy) & 0.416 & 0.584 & 0.561 & 0.416\\
        \midrule
        IGEBM (Young) & 0.506 & - & - & - \\
        IGEBM (Young  $\&$ Female) & 0.367 & 0.160 & - & - \\
        IGEBM (Young  $\&$ Female $\&$ Smiling) &  0.604 & 0.648 & 0.625 & -\\
        IGEBM  (Young  $\&$ Female $\&$ Smiling $\&$ Wavy) & 0.550   & 0.545 & 0.445 & 0.781   \\
        \midrule
        Ours (Young)   & \textbf{0.847} & - & - & - \\
        Ours (Young  $\&$ Female) & \textbf{0.770} & \textbf{0.583} &- &- \\
        Ours (Young  $\&$ Female $\&$ Smiling) & \textbf{0.906} & \textbf{0.718} & \textbf{0.968} & - \\
        Ours  (Young  $\&$ Female $\&$ Smiling $\&$ Wavy) & \textbf{0.922} & \textbf{0.625} & \textbf{0.843} & \textbf{0.906}\\

        \bottomrule
    \end{tabular}
    }
    \label{tbl:comp_table}
    \end{center}
    \vspace{-10pt}
\end{table}
\begin{figure}[t]
    \centering
    \includegraphics[width=\linewidth]{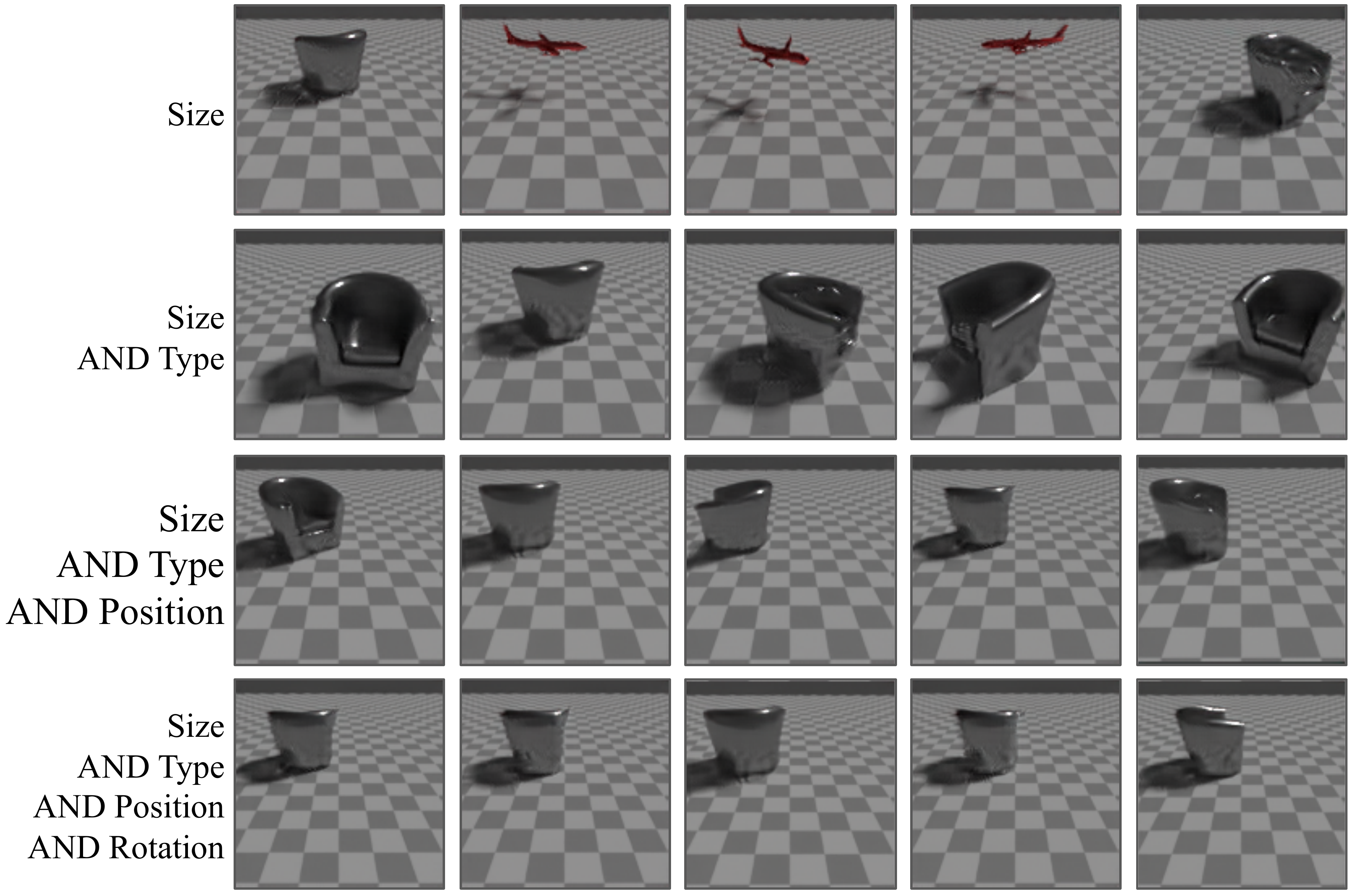}
    \vspace{-10pt}
    \caption{\small Examples of EBM compositional generations across different object attributes. Our model is successfully able to compose attributes and is able to construct high resolution, globally coherent compositional renderings, including fine detail such as lighting and reflections.}
    \label{fig:qual_cube}
    \vspace{-15pt}
\end{figure}

\begin{figure*}
\centering
\includegraphics[width=1\linewidth]{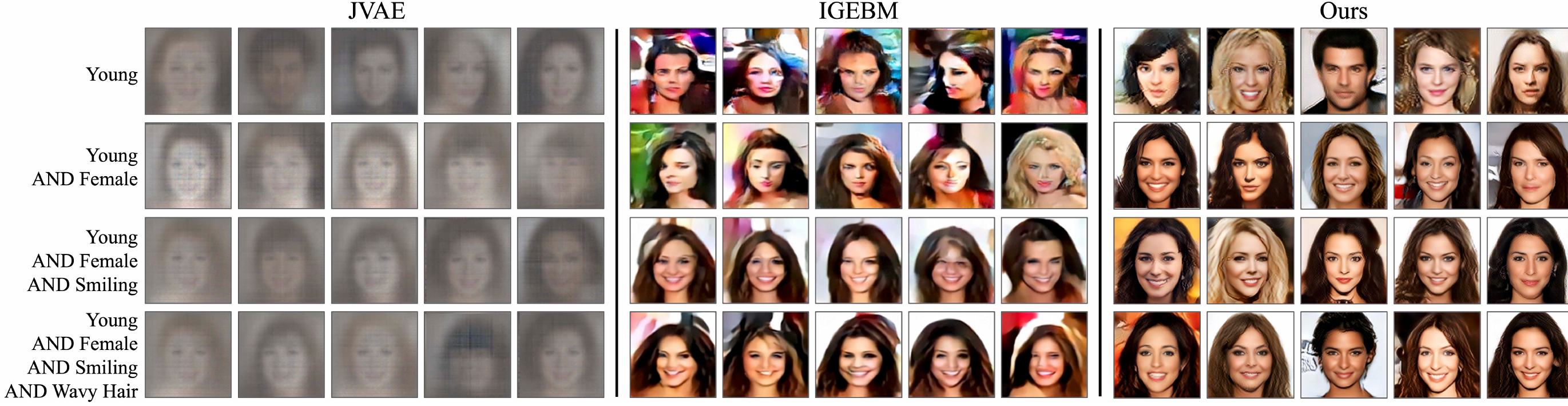}
\vspace{-20pt}
\caption{\small{Qualitative comparisons of compositionality on CelebA-HQ faces. Our approach generates much more realistic looking faces than IGEBM \citep{du2019implicit} and JVAE \citep{Vedantam2018Generative}, with each conditioned attribute.}}
\label{fig:comp_qual}
\vspace{-10pt}
\end{figure*}

\subsection{Compositionality}
\label{sect:compositionality}
Energy-Based Models (EBMs) have the ability to \textit{compose} with other models at generation time  \citep{hinton1999products, haarnoja2017reinforcement, du2020compositional}. We investigate to what extent EBMs trained under our new framework can also exhibit compositionality. See \citep{du2020compositional} for a discussion of various compositional operators and applications in EBMs. 
In particular, we train \textit{independent} EBMs $E(\vx|c_1)$, $E(\vx|c_2)$, $E(\vx|c_3)$, that learn conditional generative distribution over concept factors $c$ such as facial expression. We test whether we can compose independent energy functions together to generate images with each concept factor simultaneously. We test compositions on the CelebA-HQ dataset, where we train separate EBMs on face attributes of age, gender, smiling, and wavy hair and on a rendered Blender dataset where we train separate EBMs on object attributes of size, position, rotation, and identity.

\vspace{-3pt}
\textbf{Qualitative Results.} We present qualitative results on  compositions of energy functions on CelebA-HQ in \fig{fig:comp_qual}. We consider compositional generation on the factors of young, female, smiling, and wavy hair. Compared to baselines, our approach is able to successfully generate images with each of conditioned factor, with faces being significantly higher resolution than baselines. In \fig{fig:qual_cube}, we further consider compositions of energy functions over object attributes. We consider compositional generation on factors of size, type, position, and rotation. Again, we find that each associated image generation exhibits the corresponding conditioned attribute. Images  are further visually consistent in terms of lighting, shadows and reflections. We note that different from most past work, generations of these combinations of different factors are \textit{only} specified at generation time, with models being trained \textit{independently}. Our results indicate that our framework for training EBMs is a promising direction for high resolution compositional visual generation.

\vspace{-3pt}
\textbf{Quantitative Comparison.} We quantitatively compare compositional generations of our model with IGEBM and JVAE \citep{Vedantam2018Generative} models  on the CelebA-HQ dataset. We assess the compositional generation accuracy of different models by measuring the accuracy with which a pretrained ResNet18 classifier can recover the underlying conditioned attributes. In \tbl{tbl:comp_table}, we find that our model has significantly higher attribute recovery compared to baselines across all compositions of attributes.
\section{Related Work}
Our work is related to a large, growing body of work on different approaches for training EBMs. Our approach is based on contrastive divergence \citep{Hinton2002Training}, where an energy function is trained to contrast negative samples drawn from a model distribution and from real data. In recent years, such approaches have been applied to the image domain \citep{Salakhutdinov2009Deep, xie2016theory, gao2018learning, du2019implicit}. \citep{gao2018learning} also proposes a multiscale approach towards generating images from EBMs, but different from our work, uses each sub-scale EBM to initialize the generation of the next EBM, while we jointly sample across each resolution. Our work builds on existing works, and aims to provide improvements in generation and stability.

A difficulty with contrastive divergence training is negative sample generation. To sidestep this issue, a separate line of work utilizes an auxiliary network to amortize the negative portions of the sampling procedure \citep{kim2016deep, kumar2019maximum, han2019divergence, xie2018cooperative, song2018learning, dai2019exponential, che2020gan,  grathwohl2020vera, arbel2020ebm}. One line of work  \citep{kim2016deep, kumar2019maximum, song2018learning}, utilizes a separate generator network for negative image sample generations.  In contrast,  \citep{xie2018cooperative}, utilizes a generator to warm start generations for negative samples and \citep{han2019divergence} minimizes a divergence triangle between three models.  While such approaches enable better qualitative generation, they also lose some of the flexibility of the EBM formulation. For example, separate energy models can no longer be composed together for generation.

In addition, other approaches towards training EBMs seek instead to investigate separate objectives to train the EBM.  One such approach is score matching, where the gradients of an energy function are trained to match the gradients of real data \citep{hyvarinen2005estimation, song2019generative}, or with a related denoising  \citep{sohl2015deep, saremi2018deep, ho2020denoising} approach. Additional objectives include noise contrastive estimation \citep{gao2020flow}, learned Steins discrepancy \citep{grathwohl2020cutting}, and learned F divergences \citep{yu2020training}. 

Most prior work in contrastive divergence has ignored the KL term \citep{hinton1999products, Salakhutdinov2009Deep}. A notable exception is \citep{ruiz2019contrastive}, which obtains a similar KL divergence term to ours. \citet{ruiz2019contrastive} use a high variance REINFORCE estimator to estimate the gradient of the KL term, while our approach relies on auto-differentiation and nearest neighbor entropy estimators. Differentiation through model generation procedures has previously been explored in other models \citep{Finn2017Meta, metz2016unrolled}. Other related entropy estimators include those based on Stein's identity \citep{liu2017stein} and MINE \citep{belghazi2018mine}. In contrast to these approaches, our entropy estimator relies only on nearest neighbor calculation, and does not require the training of an independent neural network.

\vspace{-5pt}
\section{Conclusion}
\vspace{-5pt}
We propose a simple and general framework for improving generation and ease of training of EBMs. We show that the framework enables high resolution compositional image generation and out-of-distribution robustness. In the future, we are interested in further computational scaling of our framework and its applications to additional domains such as text, video, and reasoning. 

\section{Acknowledgements}

We would like to thank Jasha Sohl Dickstein, Bo Dai, Rif Sauros, Simon Osindero, Alex Alemi and anonymous reviewers for there helpful feedback on initial versions of the manuscript. Yilun Du is supported by a NSF GFRP fellowship. This work is in part supported by ONR MURI N00014-18-1-2846 and IBM Thomas J. Watson Research Center CW3031624.

\bibliography{iclr2021_conference, reference}

\begin{thebibliography}{67}
\providecommand{\natexlab}[1]{#1}
\providecommand{\url}[1]{\texttt{#1}}
\expandafter\ifx\csname urlstyle\endcsname\relax
  \providecommand{\doi}[1]{doi: #1}\else
  \providecommand{\doi}{doi: \begingroup \urlstyle{rm}\Url}\fi

\bibitem[Arbel et~al.(2020)Arbel, Zhou, and Gretton]{arbel2020ebm}
Michael Arbel, Liang Zhou, and Arthur Gretton.
\newblock Generalized energy based models.
\newblock \emph{arXiv preprint arXiv.org/abs/2003.05033}, 2020.

\bibitem[Bartunov et~al.(2019)Bartunov, Rae, Osindero, and
  Lillicrap]{bartunov2019meta}
Sergey Bartunov, Jack~W Rae, Simon Osindero, and Timothy~P Lillicrap.
\newblock Meta-learning deep energy-based memory models.
\newblock \emph{arXiv preprint arXiv:1910.02720}, 2019.

\bibitem[Beirlant et~al.(1997)Beirlant, Dudewicz, Gyor, and
  Meulen]{beirlant1997nonparametric}
Jan Beirlant, E.~Dudewicz, L.~Gyor, and E.C. Meulen.
\newblock Nonparametric entropy estimation: An overview.
\newblock \emph{International Journal of Mathematical and Statistical
  Sciences}, 6, 01 1997.

\bibitem[Belghazi et~al.(2018)Belghazi, Baratin, Rajeswar, Ozair, Bengio,
  Courville, and Hjelm]{belghazi2018mine}
Mohamed~Ishmael Belghazi, Aristide Baratin, Sai Rajeswar, Sherjil Ozair, Yoshua
  Bengio, Aaron Courville, and R~Devon Hjelm.
\newblock Mine: mutual information neural estimation.
\newblock \emph{arXiv preprint arXiv:1801.04062}, 2018.

\bibitem[Che et~al.(2020)Che, Zhang, Sohl-Dickstein, Larochelle, Paull, Cao,
  and Bengio]{che2020gan}
Tong Che, Ruixiang Zhang, Jascha Sohl-Dickstein, Hugo Larochelle, Liam Paull,
  Yuan Cao, and Yoshua Bengio.
\newblock Your gan is secretly an energy-based model and you should use
  discriminator driven latent sampling.
\newblock \emph{arXiv preprint arXiv.org/abs/2003.06060}, 2020.

\bibitem[Chen et~al.(2019)Chen, Zhai, Ritter, Lucic, and Houlsby]{chen2019self}
Ting Chen, Xiaohua Zhai, Marvin Ritter, Mario Lucic, and Neil Houlsby.
\newblock Self-supervised gans via auxiliary rotation loss.
\newblock In \emph{Proceedings of the IEEE Conference on Computer Vision and
  Pattern Recognition}, pp.\  12154--12163, 2019.

\bibitem[Chen et~al.(2020)Chen, Kornblith, Norouzi, and Hinton]{chen2020simple}
Ting Chen, Simon Kornblith, Mohammad Norouzi, and Geoffrey Hinton.
\newblock A simple framework for contrastive learning of visual
  representations.
\newblock \emph{arXiv preprint arXiv:2002.05709}, 2020.

\bibitem[Dai et~al.(2019)Dai, Liu, Dai, He, Gretton, Song, and
  Schuurmans]{dai2019exponential}
Bo~Dai, Zhen Liu, Hanjun Dai, Niao He, Arthur Gretton, Le~Song, and Dale
  Schuurmans.
\newblock Exponential family estimation via adversarial dynamics embedding.
\newblock In \emph{Advances in Neural Information Processing Systems}, pp.\
  10979--10990, 2019.

\bibitem[Deng et~al.(2009)Deng, Dong, Socher, Li, Li, and
  Fei-Fei]{Deng2009Imagenet}
Jia Deng, Wei Dong, Richard Socher, Li-Jia Li, Kai Li, and Li~Fei-Fei.
\newblock Imagenet: A large-scale hierarchical image database.
\newblock In \emph{CVPR}, 2009.

\bibitem[Deng et~al.(2020)Deng, Bakhtin, Ott, Szlam, and
  Ranzato]{deng2020residual}
Yuntian Deng, Anton Bakhtin, Myle Ott, Arthur Szlam, and Marc'Aurelio Ranzato.
\newblock Residual energy-based models for text generation.
\newblock \emph{arXiv preprint arXiv:2004.11714}, 2020.

\bibitem[Du \& Mordatch(2019)Du and Mordatch]{du2019implicit}
Yilun Du and Igor Mordatch.
\newblock Implicit generation and generalization in energy-based models.
\newblock \emph{arXiv preprint arXiv:1903.08689}, 2019.

\bibitem[Du et~al.(2019)Du, Lin, and Mordatch]{du2019model}
Yilun Du, Toru Lin, and Igor Mordatch.
\newblock Model based planning with energy based models.
\newblock \emph{arXiv preprint arXiv:1909.06878}, 2019.

\bibitem[Du et~al.(2020{\natexlab{a}})Du, Li, and
  Mordatch]{du2020compositional}
Yilun Du, Shuang Li, and Igor Mordatch.
\newblock Compositional visual generation with energy based models.
\newblock In \emph{Advances in Neural Information Processing Systems},
  2020{\natexlab{a}}.

\bibitem[Du et~al.(2020{\natexlab{b}})Du, Meier, Ma, Fergus, and
  Rives]{du2020energy}
Yilun Du, Joshua Meier, Jerry Ma, Rob Fergus, and Alexander Rives.
\newblock Energy-based models for atomic-resolution protein conformations.
\newblock \emph{arXiv preprint arXiv:2004.13167}, 2020{\natexlab{b}}.

\bibitem[Finn \& Levine(2017)Finn and Levine]{Finn2017Meta}
Chelsea Finn and Sergey Levine.
\newblock Meta-learning and universality: Deep representations and gradient
  descent can approximate any learning algorithm.
\newblock \emph{arXiv:1710.11622}, 2017.

\bibitem[Gao et~al.(2018)Gao, Lu, Zhou, Zhu, and Nian~Wu]{gao2018learning}
Ruiqi Gao, Yang Lu, Junpei Zhou, Song-Chun Zhu, and Ying Nian~Wu.
\newblock Learning generative convnets via multi-grid modeling and sampling.
\newblock In \emph{Proceedings of the IEEE Conference on Computer Vision and
  Pattern Recognition}, pp.\  9155--9164, 2018.

\bibitem[Gao et~al.(2020)Gao, Nijkamp, Kingma, Xu, Dai, and Wu]{gao2020flow}
Ruiqi Gao, Erik Nijkamp, Diederik~P Kingma, Zhen Xu, Andrew~M Dai, and
  Ying~Nian Wu.
\newblock Flow contrastive estimation of energy-based models.
\newblock In \emph{Proceedings of the IEEE/CVF Conference on Computer Vision
  and Pattern Recognition}, pp.\  7518--7528, 2020.

\bibitem[Grathwohl et~al.(2019)Grathwohl, Wang, Jacobsen, Duvenaud, Norouzi,
  and Swersky]{grathwohl2019your}
Will Grathwohl, Kuan-Chieh Wang, J{\"o}rn-Henrik Jacobsen, David Duvenaud,
  Mohammad Norouzi, and Kevin Swersky.
\newblock Your classifier is secretly an energy based model and you should
  treat it like one.
\newblock \emph{arXiv preprint arXiv:1912.03263}, 2019.

\bibitem[Grathwohl et~al.(2020{\natexlab{a}})Grathwohl, Kelly, Hashemi,
  Norouzi, Swersky, and Duvenaud]{grathwohl2020vera}
Will Grathwohl, Jacob Kelly, Milad Hashemi, Mohammad Norouzi, Kevin Swersky,
  and David Duvenaud.
\newblock No mcmc for me: Amortized sampling for fast and stable training of
  energy-based models.
\newblock \emph{arXiv preprint arxiv.org/abs/2010.04230}, 2020{\natexlab{a}}.

\bibitem[Grathwohl et~al.(2020{\natexlab{b}})Grathwohl, Wang, Jacobsen,
  Duvenaud, and Zemel]{grathwohl2020cutting}
Will Grathwohl, Kuan-Chieh Wang, Jorn-Henrik Jacobsen, David Duvenaud, and
  Richard Zemel.
\newblock Cutting out the middle-man: Training and evaluating energy-based
  models without sampling.
\newblock \emph{arXiv preprint arXiv:2002.05616}, 2020{\natexlab{b}}.

\bibitem[Gulrajani et~al.(2017)Gulrajani, Ahmed, Arjovsky, Dumoulin, and
  Courville]{Gulrajani2017Improved}
Ishaan Gulrajani, Faruk Ahmed, Martin Arjovsky, Vincent Dumoulin, and Aaron
  Courville.
\newblock Improved training of wasserstein gans.
\newblock In \emph{NIPS}, 2017.

\bibitem[Haarnoja et~al.(2017)Haarnoja, Tang, Abbeel, and
  Levine]{haarnoja2017reinforcement}
Tuomas Haarnoja, Haoran Tang, Pieter Abbeel, and Sergey Levine.
\newblock Reinforcement learning with deep energy-based policies.
\newblock \emph{arXiv preprint arXiv:1702.08165}, 2017.

\bibitem[Han et~al.(2019)Han, Nijkamp, Fang, Hill, Zhu, and
  Wu]{han2019divergence}
Tian Han, Erik Nijkamp, Xiaolin Fang, Mitch Hill, Song-Chun Zhu, and Ying~Nian
  Wu.
\newblock Divergence triangle for joint training of generator model,
  energy-based model, and inferential model.
\newblock In \emph{Proceedings of the IEEE Conference on Computer Vision and
  Pattern Recognition}, pp.\  8670--8679, 2019.

\bibitem[Hendrycks \& Gimpel(2016)Hendrycks and Gimpel]{hendrycks2016baseline}
Dan Hendrycks and Kevin Gimpel.
\newblock A baseline for detecting misclassified and out-of-distribution
  examples in neural networks.
\newblock \emph{arXiv preprint arXiv:1610.02136}, 2016.

\bibitem[Heusel et~al.(2017)Heusel, Ramsauer, Unterthiner, Nessler, and
  Hochreiter]{heusel2017gans}
Martin Heusel, Hubert Ramsauer, Thomas Unterthiner, Bernhard Nessler, and Sepp
  Hochreiter.
\newblock Gans trained by a two time-scale update rule converge to a local nash
  equilibrium.
\newblock In \emph{Advances in Neural Information Processing Systems}, pp.\
  6626--6637, 2017.

\bibitem[Hinton(1999)]{hinton1999products}
Geoffrey~E Hinton.
\newblock Products of experts.
\newblock \emph{International Conference on Artificial Neural Networks}, 1999.

\bibitem[Hinton(2002)]{Hinton2002Training}
Geoffrey~E Hinton.
\newblock Training products of experts by minimizing contrastive divergence.
\newblock \emph{Neural Comput.}, 14\penalty0 (8):\penalty0 1771--1800, 2002.

\bibitem[Ho et~al.(2020)Ho, Jain, and Abbeel]{ho2020denoising}
Jonathan Ho, Ajay Jain, and Pieter Abbeel.
\newblock Denoising diffusion probabilistic models.
\newblock \emph{arXiv preprint arXiv:2006.11239}, 2020.

\bibitem[Hyv{\"a}rinen(2005)]{hyvarinen2005estimation}
Aapo Hyv{\"a}rinen.
\newblock Estimation of non-normalized statistical models by score matching.
\newblock \emph{Journal of Machine Learning Research}, 6\penalty0
  (Apr):\penalty0 695--709, 2005.

\bibitem[Ingraham et~al.()Ingraham, Riesselman, Sander, and
  Marks]{ingraham2019learning}
John Ingraham, Adam Riesselman, Chris Sander, and Debora Marks.
\newblock Learning protein structure with a differentiable simulator.

\bibitem[Karras et~al.(2017)Karras, Aila, Laine, and
  Lehtinen]{Karras2017Progressive}
Tero Karras, Timo Aila, Samuli Laine, and Jaakko Lehtinen.
\newblock Progressive growing of gans for improved quality, stability, and
  variation.
\newblock In \emph{ICLR}, 2017.

\bibitem[Kim \& Bengio(2016)Kim and Bengio]{kim2016deep}
Taesup Kim and Yoshua Bengio.
\newblock Deep directed generative models with energy-based probability
  estimation.
\newblock \emph{arXiv preprint arXiv:1606.03439}, 2016.

\bibitem[Kingma \& Ba(2015)Kingma and Ba]{Kingma2015Adam}
Diederik~P. Kingma and Jimmy Ba.
\newblock Adam: A method for stochastic optimization.
\newblock In \emph{ICLR}, 2015.

\bibitem[Kozachenko \& Leonenko(1987)Kozachenko and
  Leonenko]{kozachenko1987sample}
LF~Kozachenko and Nikolai~N Leonenko.
\newblock Sample estimate of the entropy of a random vector.
\newblock \emph{Problemy Peredachi Informatsii}, 23\penalty0 (2):\penalty0
  9--16, 1987.

\bibitem[Kumar et~al.(2019)Kumar, Goyal, Courville, and
  Bengio]{kumar2019maximum}
Rithesh Kumar, Anirudh Goyal, Aaron Courville, and Yoshua Bengio.
\newblock Maximum entropy generators for energy-based models.
\newblock \emph{arXiv preprint arXiv:1901.08508}, 2019.

\bibitem[Lee et~al.(2018)Lee, Xu, Fan, and Tu]{lee2018wasserstein}
Kwonjoon Lee, Weijian Xu, Fan Fan, and Zhuowen Tu.
\newblock Wasserstein introspective neural networks.
\newblock In \emph{Proceedings of the IEEE Conference on Computer Vision and
  Pattern Recognition}, pp.\  3702--3711, 2018.

\bibitem[Li et~al.(2020)Li, Du, van~de Ven, and Mordatch]{li2020energy}
Shuang Li, Yilun Du, Gido~M van~de Ven, and Igor Mordatch.
\newblock Energy-based models for continual learning.
\newblock \emph{arXiv preprint arXiv:2011.12216}, 2020.

\bibitem[Liu \& Wang(2017)Liu and Wang]{liu2017learning}
Qiang Liu and Dilin Wang.
\newblock Learning deep energy models: Contrastive divergence vs. amortized
  mle.
\newblock \emph{arXiv preprint arXiv:1707.00797}, 2017.

\bibitem[Liu et~al.(2017)Liu, Ramachandran, Liu, and Peng]{liu2017stein}
Yang Liu, Prajit Ramachandran, Qiang Liu, and Jian Peng.
\newblock Stein variational policy gradient.
\newblock \emph{arXiv preprint arXiv:1704.02399}, 2017.

\bibitem[Lyu(2011)]{lyu2011unifying}
Siwei Lyu.
\newblock Unifying non-maximum likelihood learning objectives with minimum kl
  contraction.
\newblock In \emph{Advances in Neural Information Processing Systems}, pp.\
  64--72, 2011.

\bibitem[Metz et~al.(2016)Metz, Poole, Pfau, and
  Sohl-Dickstein]{metz2016unrolled}
Luke Metz, Ben Poole, David Pfau, and Jascha Sohl-Dickstein.
\newblock Unrolled generative adversarial networks.
\newblock 11 2016.

\bibitem[Miyato et~al.(2018)Miyato, Kataoka, Koyama, and
  Yoshida]{miyato2018spectral}
Takeru Miyato, Toshiki Kataoka, Masanori Koyama, and Yuichi Yoshida.
\newblock Spectral normalization for generative adversarial networks.
\newblock \emph{arXiv preprint arXiv:1802.05957}, 2018.

\bibitem[Neal(2011)]{neal2011mcmc}
Radford~M Neal.
\newblock Mcmc using hamiltonian dynamics.
\newblock \emph{Handbook of Markov Chain Monte Carlo}, 2\penalty0 (11), 2011.

\bibitem[Nijkamp et~al.(2019{\natexlab{a}})Nijkamp, Hill, Han, Zhu, and
  Wu]{nijkamp2019anatomy}
Erik Nijkamp, Mitch Hill, Tian Han, Song-Chun Zhu, and Ying~Nian Wu.
\newblock On the anatomy of mcmc-based maximum likelihood learning of
  energy-based models.
\newblock \emph{arXiv preprint arXiv:1903.12370}, 2019{\natexlab{a}}.

\bibitem[Nijkamp et~al.(2019{\natexlab{b}})Nijkamp, Hill, Zhu, and
  Wu]{nijkamp2019learning}
Erik Nijkamp, Mitch Hill, Song-Chun Zhu, and Ying~Nian Wu.
\newblock Learning non-convergent non-persistent short-run mcmc toward
  energy-based model.
\newblock In \emph{Advances in Neural Information Processing Systems}, pp.\
  5232--5242, 2019{\natexlab{b}}.

\bibitem[Ostrovski et~al.(2018)Ostrovski, Dabney, and
  Munos]{ostrovski2018autoregressive}
Georg Ostrovski, Will Dabney, and R{\'e}mi Munos.
\newblock Autoregressive quantile networks for generative modeling.
\newblock \emph{arXiv preprint arXiv:1806.05575}, 2018.

\bibitem[Radford et~al.(2016)Radford, Metz, and
  Chintala]{Radford2016Unsupervised}
Alec Radford, Luke Metz, and Soumith Chintala.
\newblock Unsupervised representation learning with deep convolutional
  generative adversarial networks.
\newblock In \emph{ICLR}, 2016.

\bibitem[Ramachandran et~al.(2018)Ramachandran, Zoph, and
  Le]{ramachandran2018searching}
Prajit Ramachandran, Barret Zoph, and Quoc~V. Le.
\newblock Searching for activation functions, 2018.
\newblock URL \url{https://openreview.net/forum?id=SkBYYyZRZ}.

\bibitem[Ruiz \& Titsias(2019)Ruiz and Titsias]{ruiz2019contrastive}
Francisco~JR Ruiz and Michalis~K Titsias.
\newblock A contrastive divergence for combining variational inference and
  mcmc.
\newblock \emph{arXiv preprint arXiv:1905.04062}, 2019.

\bibitem[Salakhutdinov \& Hinton(2009)Salakhutdinov and
  Hinton]{Salakhutdinov2009Deep}
Ruslan Salakhutdinov and Geoffrey~E. Hinton.
\newblock Deep boltzmann machines.
\newblock In David A.~Van Dyk and Max Welling (eds.), \emph{AISTATS}, volume~5
  of \emph{{JMLR} Proceedings}, pp.\  448--455. JMLR.org, 2009.
\newblock URL
  \url{http://www.jmlr.org/proceedings/papers/v5/salakhutdinov09a.html}.

\bibitem[Salimans et~al.(2016)Salimans, Goodfellow, Zaremba, Cheung, Radford,
  and Chen]{Salimans2016Improved}
Tim Salimans, Ian Goodfellow, Wojciech Zaremba, Vicki Cheung, Alec Radford, and
  Xi~Chen.
\newblock Improved techniques for training gans.
\newblock In \emph{NIPS}, 2016.

\bibitem[Saremi et~al.(2018)Saremi, Mehrjou, Sch{\"o}lkopf, and
  Hyv{\"a}rinen]{saremi2018deep}
Saeed Saremi, Arash Mehrjou, Bernhard Sch{\"o}lkopf, and Aapo Hyv{\"a}rinen.
\newblock Deep energy estimator networks.
\newblock \emph{arXiv preprint arXiv:1805.08306}, 2018.

\bibitem[Scellier \& Bengio(2017)Scellier and Bengio]{scellier2017equilibrium}
Benjamin Scellier and Yoshua Bengio.
\newblock Equilibrium propagation: Bridging the gap between energy-based models
  and backpropagation.
\newblock \emph{Frontiers in computational neuroscience}, 11:\penalty0 24,
  2017.

\bibitem[Sohl-Dickstein et~al.(2015)Sohl-Dickstein, Weiss, Maheswaranathan, and
  Ganguli]{sohl2015deep}
Jascha Sohl-Dickstein, Eric~A Weiss, Niru Maheswaranathan, and Surya Ganguli.
\newblock Deep unsupervised learning using nonequilibrium thermodynamics.
\newblock \emph{arXiv preprint arXiv:1503.03585}, 2015.

\bibitem[Song \& Ermon(2019)Song and Ermon]{song2019generative}
Yang Song and Stefano Ermon.
\newblock Generative modeling by estimating gradients of the data distribution.
\newblock In \emph{Advances in Neural Information Processing Systems}, pp.\
  11918--11930, 2019.

\bibitem[Song \& Ou(2018)Song and Ou]{song2018learning}
Yunfu Song and Zhijian Ou.
\newblock Learning neural random fields with inclusive auxiliary generators.
\newblock \emph{arXiv preprint arXiv:1806.00271}, 2018.

\bibitem[Tsybakov \& Van~der Meulen(1996)Tsybakov and Van~der
  Meulen]{tsybakov1996root}
Alexandre~B Tsybakov and EC~Van~der Meulen.
\newblock Root-n consistent estimators of entropy for densities with unbounded
  support.
\newblock \emph{Scandinavian Journal of Statistics}, pp.\  75--83, 1996.

\bibitem[Vahdat et~al.(2020)Vahdat, Andriyash, and
  Macready]{pmlr-v119-vahdat20a}
Arash Vahdat, Evgeny Andriyash, and William Macready.
\newblock Undirected graphical models as approximate posteriors.
\newblock In Hal~Daumé III and Aarti Singh (eds.), \emph{Proceedings of the
  37th International Conference on Machine Learning}, volume 119 of
  \emph{Proceedings of Machine Learning Research}, pp.\  9680--9689. PMLR,
  13--18 Jul 2020.
\newblock URL \url{http://proceedings.mlr.press/v119/vahdat20a.html}.

\bibitem[van~den Oord et~al.(2016)van~den Oord, Kalchbrenner, Espeholt,
  Vinyals, Graves, et~al.]{Oord2016Conditional}
Aaron van~den Oord, Nal Kalchbrenner, Lasse Espeholt, Oriol Vinyals, Alex
  Graves, et~al.
\newblock Conditional image generation with pixelcnn decoders.
\newblock In \emph{NIPS}, 2016.

\bibitem[Van~Oord et~al.(2016)Van~Oord, Kalchbrenner, and
  Kavukcuoglu]{VanOord2016Pixel}
Aaron Van~Oord, Nal Kalchbrenner, and Koray Kavukcuoglu.
\newblock Pixel recurrent neural networks.
\newblock In \emph{ICML}, 2016.

\bibitem[Vedantam et~al.(2018)Vedantam, Fischer, Huang, and
  Murphy]{Vedantam2018Generative}
Ramakrishna Vedantam, Ian Fischer, Jonathan Huang, and Kevin Murphy.
\newblock Generative models of visually grounded imagination.
\newblock In \emph{ICLR}, 2018.

\bibitem[Wu \& He(2018)Wu and He]{Wu2018Group}
Yuxin Wu and Kaiming He.
\newblock Group normalization.
\newblock \emph{arXiv:1803.08494}, 2018.

\bibitem[Xie et~al.(2016)Xie, Lu, Zhu, and Wu]{xie2016theory}
Jianwen Xie, Yang Lu, Song-Chun Zhu, and Yingnian Wu.
\newblock A theory of generative convnet.
\newblock In \emph{International Conference on Machine Learning}, pp.\
  2635--2644, 2016.

\bibitem[Xie et~al.(2017)Xie, Zhu, and Nian~Wu]{xie2018video}
Jianwen Xie, Song-Chun Zhu, and Ying Nian~Wu.
\newblock Synthesizing dynamic patterns by spatial-temporal generative convnet.
\newblock In \emph{Proceedings of the IEEE conference on computer vision and
  pattern recognition}, 2017.

\bibitem[Xie et~al.(2018{\natexlab{a}})Xie, Lu, Gao, and
  Wu]{xie2018cooperative}
Jianwen Xie, Yang Lu, Ruiqi Gao, and Ying~Nian Wu.
\newblock Cooperative learning of energy-based model and latent variable model
  via mcmc teaching.
\newblock In \emph{Thirty-Second AAAI Conference on Artificial Intelligence},
  2018{\natexlab{a}}.

\bibitem[Xie et~al.(2018{\natexlab{b}})Xie, Zheng, Gao, Wang, Zhu, and
  Nian~Wu]{xie2018learning}
Jianwen Xie, Zilong Zheng, Ruiqi Gao, Wenguan Wang, Song-Chun Zhu, and Ying
  Nian~Wu.
\newblock Learning descriptor networks for 3d shape synthesis and analysis.
\newblock In \emph{Proceedings of the IEEE conference on computer vision and
  pattern recognition}, pp.\  8629--8638, 2018{\natexlab{b}}.

\bibitem[Yu et~al.(2020)Yu, Song, Song, and Ermon]{yu2020training}
Lantao Yu, Yang Song, Jiaming Song, and Stefano Ermon.
\newblock Training deep energy-based models with f-divergence minimization.
\newblock \emph{arXiv preprint arXiv:2003.03463}, 2020.

\end{thebibliography}
\bibliographystyle{iclr2021_conference}

\newpage
\newpage
\appendix
In this supplement, we present additional image generation results in \sect{sect:image}. Next we detail experimental settings in \sect{sect:method}.  We provide derivations of  gradients of $\mathcal{L}_{\text{CD}}$ and $\mathcal{L}_{\text{KL}}$ and show there equivalence to the original contrastive divergence objective in \sect{sect:objective}.  Finally we provide additional analysis of our method in \sect{sect:analysis}.

\section{More Image Results}
\label{sect:image}
\subsection{Nearest Neighbor Generations}

\begin{figure}[h]
    \centering
    \includegraphics[width=1\linewidth]{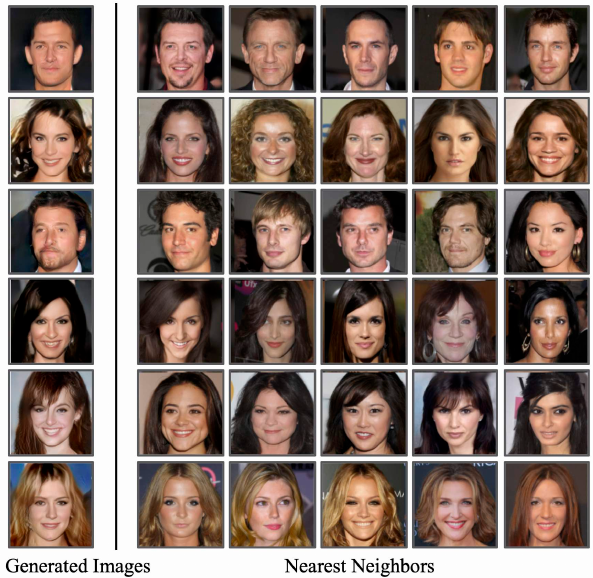}
    \vspace{-5pt}
    \caption{\small Nearest neighbors in the L2 space of generated images in CelebA-HQ 128x128.}
    \label{fig:celeba_nn}
\end{figure}

We present L2 nearest neighbors in CelebA-HQ training dataset of unconditional image samples from our trained EBM in \fig{fig:celeba_nn}. We find that our approach generates images distinct from the training set.

\subsection{Additional Quantitative Results}

We further quantitatively compare our generations with those of SNGAN on LSUN 128x128 bedroom scenes. We find that an SNGAN model trained on LSUN 128x128 bedroom scenes obtains an FID of 64.05 compared to our approach, which obtains an FID of 33.46. To report SNGAN scores, we re-implemented the SNGAN model using the default hyper parameters to train models on ImageNet 128x128.

\subsection{Additional Qualitative Images}

\begin{figure}
    \centering
    \includegraphics[width=1.0\linewidth]{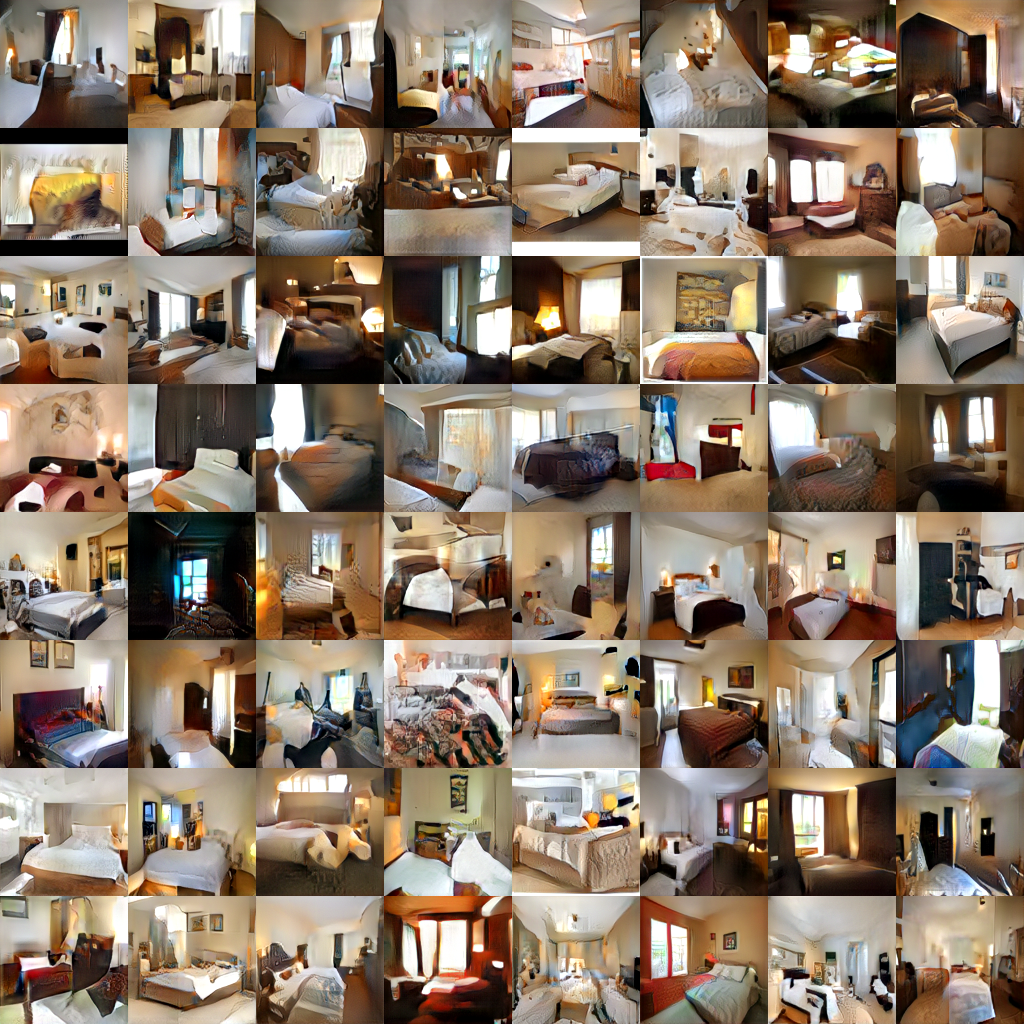}
    \vspace{-10pt}
    \caption{\small Randomly selected unconditional LSUN bed 128x128 samples from our trained EBM.} 
    \label{fig:lsun_uncond}
\end{figure}

\begin{figure}
\vspace{-12pt}
    \centering
    \includegraphics[width=1.0\linewidth]{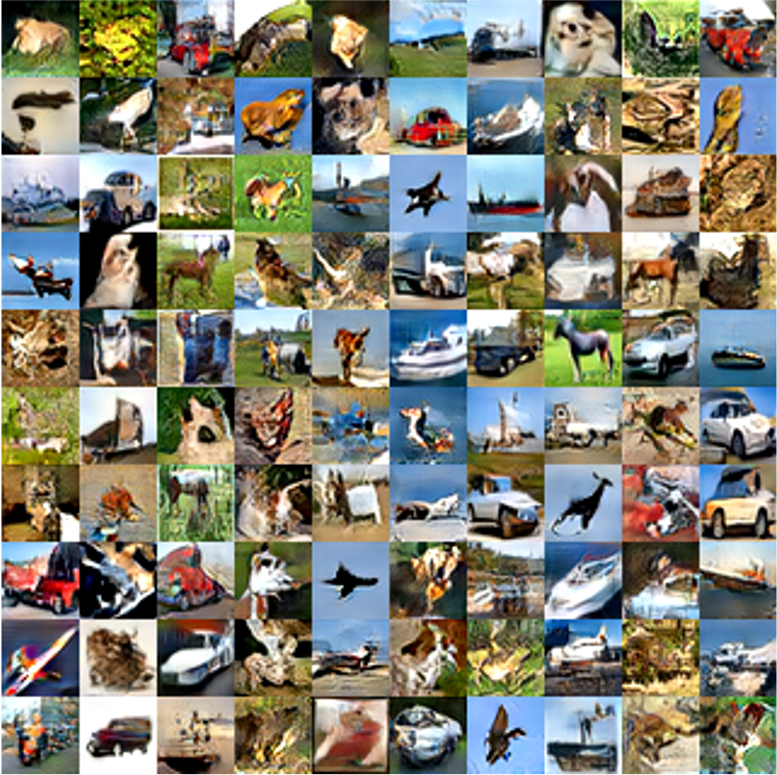}
    \vspace{-10pt}
    \caption{\small Randomly selected unconditional CIFAR-10 samples from our trained EBM.} 
    \label{fig:cifar_uncond}
\end{figure}

\begin{figure}
    \centering
    \includegraphics[width=1.0\linewidth]{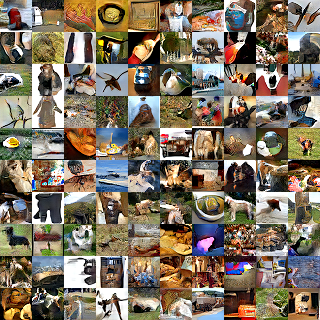}
    \vspace{-10pt}
    \caption{\small Randomly selected unconditional ImageNet 32x32 samples from our trained EBM.} 
    \label{fig:imagenet_uncond}
\end{figure}

We present qualitative visualizations of unconditional samples generated from an EBM. \fig{fig:lsun_uncond} shows unconditional image generations from LSUN bedroom scenes. \fig{fig:cifar_uncond} shows unconditional image generations on the CIFAR-10 dataset. Finally, \fig{fig:imagenet_uncond} shows unconditional image generations on the ImageNet 32x32 dataset. In all three different settings, we find that our generated unconditional images are relatively globally coherent.

\section{Training Details}
\label{sect:method}
\subsection{Model Architectures}

In this part, we provide the model architectures used in  our experiments. When training multiscale energy functions, our final output energy function is the sum of energy functions applied to the full resolution image, the half resolution image, and the quarter resolution image. We use the architecture reported in \tbl{fig:cifar} for the full resolution image on CIFAR-10 and ImageNet 32x32 (used in the main paper Section 3.2 and 3.3). The model architecture used on the CelebA-HQ and LSUN datasets are reported in \tbl{fig:celeba_arch} (used in the main paper Section 3.2 and 3.4). The half-resolution models share the architecture listed in \tbl{fig:cifar}, but with the first down-sampled residual block removed. Similarly, the quarter resolution models share the architectures listed, but with the first two down-sampled residual blocks removed. We utilize group normalization \citep{Wu2018Group} inside each residual block and utilize the Swish nonlinearity \citep{ramachandran2018searching}.

\vspace{-10pt}
\subsection{Experiment Configurations For Different Datasets}

\myparagraph{CIFAR-10/ImageNet 32x32.} For CIFAR-10 and ImageNet 32x32, we use 40 steps of Langevin sampling to generate a negative sample. The Langevin sampling step size is set to be 500, with Gaussian noise of magnitude 0.001 at each iteration. The data augmentation transform consists of color augmentation of strength 1.0 from \citep{chen2020simple}, a random horizontal flip, and a image resize between 0.02 and 1.0. This is used in the main paper Section 3.2 and 3.3.

\myparagraph{CelebA/LSUN Bedroom.} For the CelebA-HQ and LSUN bed datasets, we use 40 steps of Langevin sampling to generate negative samples. The Langevin sampling step size is set to be 1000, with Gaussian noise of magnitude 0.001 applied at each iteration. The data augmentation transform consists of color augmentation of strength 1.0 from \citep{chen2020simple}, a random horizontal flip, and a image resize between 0.08 and 1.0. This is used in the main paper Section 3.2 and 3.4.

\section{Loss Gradient Derivation}
\label{sect:objective}

We show that the gradient of the contrastive divergence objective, $\mathcal{L}_{\text{CD\_Full}}$ is equivalent to that of the $\mathcal{L}_{\text{Full}} = \mathcal{L}_{\text{KL}} + \mathcal{L}_{\text{CD}}$. Recall that the contrastive divergence objective is given by
\begin{equation}
    \mathcal{L}_{\text{CD\_Full}} = \kldiv{p_D(\vx)}{p_{\theta} (\vx)} - \kldiv{q_{\theta}(\vx)}{p_{\theta}(\vx)}. 
    \label{eq:cd_sup}
\end{equation}
The gradient of the first KL term with respect to $\theta$, $\frac{\partial \kldiv{p_D(\vx)}{p_{\theta} (\vx)}}{\partial \theta}$  is
\begin{equation}
   \resizebox{0.3\hsize}{!}{$
  - \E_{p_D(\vx)}  \left [\frac{\partial E_{\theta}(\vx)}{\partial \theta} \right] 
  $}
\end{equation}
while the gradient of the second KL term with respect to $\theta$, $\frac{\kldiv{q_{\theta}(\vx)}{p_{\theta}(\vx)}}{\partial \theta}$
\begin{equation}
   \resizebox{0.8\hsize}{!}{$ 
  \frac{\partial q(\vx')}{\partial \theta} \frac{\partial \kldiv{q_{\theta}(\vx')}{p_{\theta}(\vx')}}{\partial q_{\theta}(\vx')}) - \E_{q_{\theta}(\vx')} [\frac{\partial E_{\theta}(\vx')}{\partial \theta}]  
  $}
\end{equation}
with the overall gradient being
\begin{equation}
    \label{eq:cd_obj_grad_sup}
    \resizebox{0.8\hsize}{!}{$
    \begin{split}
    \frac{\mathcal{L}_{\text{CD\_Full}}}{\partial \theta} =
    - ( \E_{p_D(\vx)} \left [\frac{\partial E_{\theta}(\vx)}{\partial \theta} \right] &- \E_{q_{\theta}(\vx')} [\frac{\partial E_{\theta}(\vx')}{\partial \theta}] \\
    &+ {\color{red} \frac{\partial q(\vx')}{\partial \theta} \frac{\partial \kldiv{q_{\theta}(\vx')}{p_{\theta}(\vx')}}{\partial q_{\theta}(\vx')})}
    \end{split}$}.
\end{equation}
We have that 
\begin{equation}
    \mathcal{L}_{\text{CD}} =  \E_{p_D(\vx)} [E_{\theta}(\vx)] - \E_{\text{stop\_grad}(q_{\theta}(\vx'))} [E_{\theta}(\vx')],
\end{equation}
with  corresponding gradients
\begin{equation}
    \resizebox{0.8\hsize}{!}{$
     \frac{\partial \mathcal{L}_{\text{CD}}}{\partial \theta} = \E_{p_D(\vx)} \left [\frac{\partial E_{\theta}(\vx)}{\partial \theta} \right] - \E_{q_{\theta}(\vx')} [\frac{\partial E_{\theta}(\vx')}{\partial \theta}]$}. 
    \label{eq:cd_loss_grad_sup}
\end{equation}
Furthermore, we have that 
\begin{equation}
    \mathcal{L}_{\text{KL}} =  \E_{q_{\theta}(\vx)} [E_{\text{stop\_grad}(\theta)}(\vx)] + \E_{q_{\theta}(\vx)}[\log (q_{\theta}(\vx))] ,
\end{equation}
can be rewritten as 
\begin{align}
    \mathcal{L}_{\text{KL}} &=  \E_{q_{\theta}(\vx)} [-\log(p_{\theta}(x))] + \E_{q_{\theta}(\vx)}[\log (q_{\theta}(\vx))]  \\
    &= \kldiv{q_\theta(\vx)}{p_{\text{stop\_gradient}(\theta)}(\vx)}.
\end{align}

The corresponding gradient of the objective is
\begin{equation}
     \frac{\partial \mathcal{L}_{\text{KL}}}{\partial \theta} =\frac{\partial q(\vx)}{\partial \theta} \frac{\partial \kldiv{q_{\theta}(\vx)}{p_{\theta}(\vx)}}{\partial q_{\theta}(\vx)}. 
    \label{eq:kl_loss_grad_sup}
\end{equation}

Thus the sum of the gradients in $\frac{\partial \mathcal{L}_{\text{CD}}}{\partial \theta}$ (\eqn{eq:cd_loss_grad_sup}) and $\frac{\partial \mathcal{L}_{\text{KL}}}{\partial \theta}$ (\eqn{eq:kl_loss_grad_sup}) is equal to the full contrastive divergence gradient $\frac{\mathcal{L}_{\text{CD\_Full}}}{\partial \theta}$ (  \eqn{eq:cd_obj_grad_sup}).

\section{Additional Analysis}
\label{sect:analysis}

\subsection{Alternative Sampling Distributions} 
\begin{figure}
    \centering
    \includegraphics[width=\linewidth]{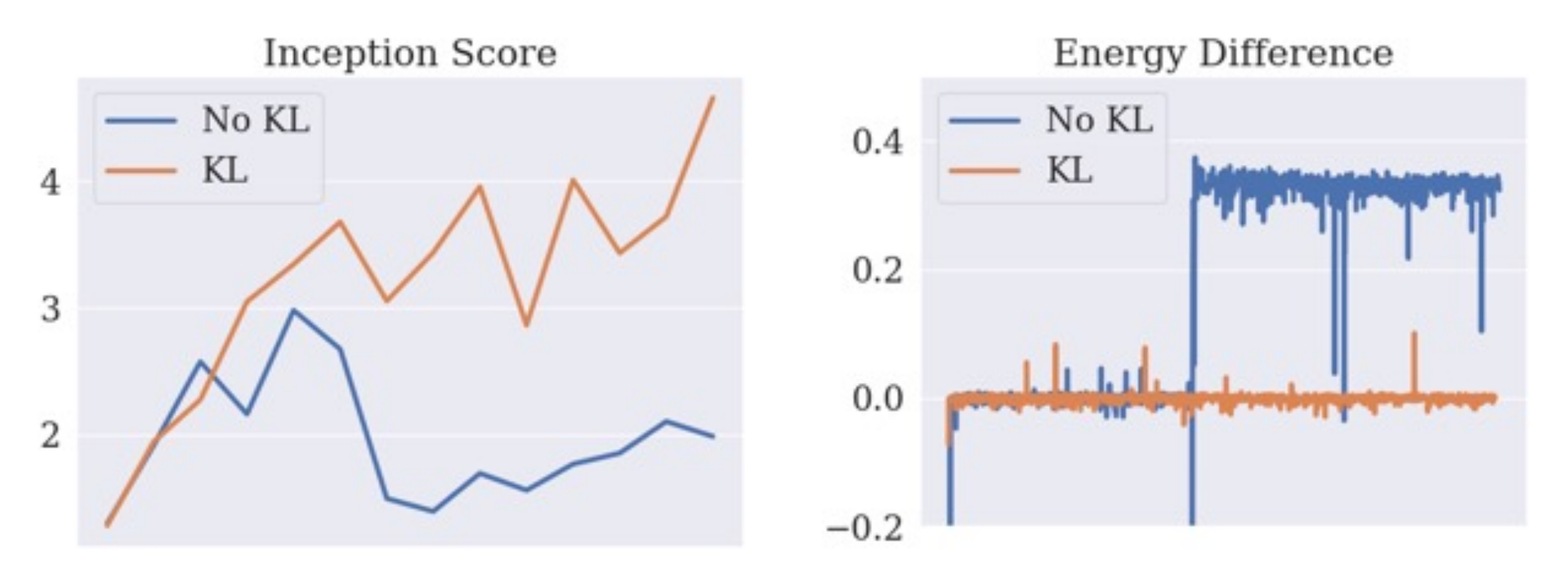}
    \vspace{-10pt}
    \caption{\small Inception Score and energy difference plots when $\mathcal{L}_{\text{KL}}$ is applied to MCMC initialized from random noise.} 
    \label{fig:noise}
    \vspace{-15pt}
\end{figure}

Instead of utilizing $q_\theta(\vx)$ as $\Pi_\theta^t(p_D(\vx))$, as noted in the method section, our approach can further maximize likelihood as long as $\kldiv{p_D(\vx)}{p_\theta(\vx)}$ is greater $\kldiv{q_\theta(\vx)}{p_\theta(\vx)}$. We test an alternative sampler $q_\theta(\vx)$ consisting of initializing Langevin dynamics from random noise in \fig{fig:noise}. We find again that our approach improves the training stability.

\subsection{Analysis of Truncated Langevin Backpropagation}
\begin{figure}
    \centering
    \includegraphics[width=1\linewidth]{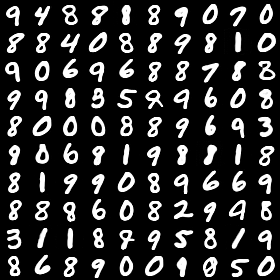}
    \caption{\small Generations on MNIST with backpropogation through 1 step of Langevin sampling. }
    \label{fig:mnist_one}
\end{figure}

\begin{figure}
    \centering
    \includegraphics[width=1\linewidth]{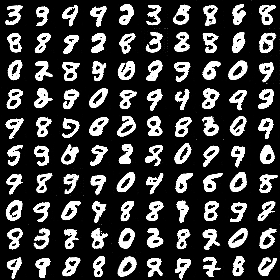}
    \caption{\small Generations on MNIST with backpropogation through all steps of Langevin sampling.}
    \label{fig:mnist_all}
\end{figure}

To better understand the training effect of $\mathcal{L}_{\text{KL}}$, we analyze the effect of truncating backpropogation through Langevin sampling.  We train two separate models on MNIST, one with backpropogation through all Langevin steps, and one with backpropogation through only the last Langevin step. We obtain an FID of 90.54 with backpropogation through only 1 step of Langevin sampling and an FID of 94.85 with backpropogation through all steps of Langevin sampling. We present illustrations of samples generated with one step in \fig{fig:mnist_one} and with all steps in \fig{fig:mnist_all}. Overall, we find little degradation in performance with the truncation of backpropogation, but note that backpropogation through all steps of sampling is over 3 times slower to train.

\subsection{Analysis of Effect of KL Loss on Mode Sampling}

\begin{figure}
    \centering
    \includegraphics[width=\linewidth]{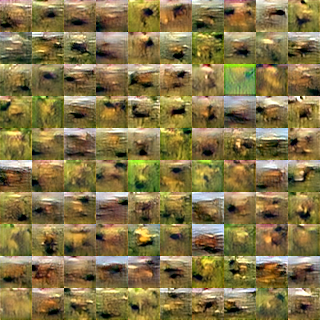}
    \vspace{-10pt}
    \caption{\small Illustration of collapsed sampling from an EBM. Sampling does not collapse with the addition of the KL loss.}
    \label{fig:collapse}
\end{figure}

We illustrate the effect of $\mathcal{L}_{\text{KL}}$ as a regularizer to prevent EBM sampling collapse. When training an EBM, $\mathcal{L}_{\text{KL}}$ serves as a repelling term encouraging MCMC samples from an EBM to both have low energy and exhibit diversity. In the absence of $\mathcal{L}_{\text{KL}}$, we find that EBM sampling always collapses and eventually always generates samples illustrated in \fig{fig:collapse}. These samples are significantly less diverse than those generated when training with $\mathcal{L}_{\text{KL}}$ ( \fig{fig:cifar}), which never suffers from sampling collapse.


\begin{figure}

\begin{minipage}{0.45\linewidth}
\captionof{table}{The model architecture used for CIFAR-10 and ImageNet-32x32 experiments.}
\centering
\begin{tabular}{c}
    \toprule
    \toprule
    3x3 conv2d, 64 \\
    \midrule
    ResBlock  64 \\
    \midrule
    ResBlock Down 64 \\
    \midrule
    ResBlock  64 \\
    \midrule
    ResBlock  Down 64 \\
    \midrule
    Self Attention 64 \\
    \midrule
    ResBlock  128 \\
    \midrule
    ResBlock  Down 128 \\
    \midrule
    ResBlock  256 \\
    \midrule
    ResBlock  Down 256 \\
    \midrule
    Global Mean Pooling \\ 
    \midrule
    Dense $\rightarrow$ 1 \\ 
    \bottomrule
\end{tabular}
\label{fig:cifar}
\end{minipage}%
\hfill
\begin{minipage}{0.45\linewidth}
\vspace{-19pt}
\captionof{table}{The model architecture used for CelebA-HQ/LSUN room experiments.}
\centering
\begin{tabular}{c}
    \toprule
    \toprule
    3x3 conv2d, 64 \\
    \midrule
    ResBlock Down 64 \\
    \midrule
    ResBlock Down 128 \\
    \midrule
    ResBlock Down 128 \\
    \midrule
    ResBlock 256 \\
    \midrule
    ResBlock Down 256 \\ 
    \midrule
    Self Attention 512 \\
    \midrule
    ResBlock  512 \\ 
    \midrule
    ResBlock  Down 512 \\ 
    \midrule
    Global mean Pooling \\
    \midrule
    Dense $\rightarrow$ 1 \\
    \bottomrule
\end{tabular}
\label{fig:celeba_arch}
\end{minipage}

\end{figure}

\end{document}